\documentclass[journal]{IEEEtran}
\usepackage{cite}
\usepackage{amsmath,amssymb,amsfonts}
\usepackage{algorithm, algorithmic}
\usepackage{graphicx}
\usepackage{textcomp}
\usepackage{xcolor}
\usepackage{hyperref}
\usepackage{siunitx}
\usepackage{subcaption}
\usepackage{float}
\usepackage{multirow}
\usepackage{caption}
\usepackage{diagbox}
\usepackage{pifont}
\usepackage{array, makecell}
\usepackage{titlesec}
\usepackage[super]{nth}
\usepackage{tabularx}

\usepackage{boldline}

\newcolumntype{C}{>{\centering\arraybackslash}X}

\newcommand{\etal}{\textit{et al}.}
\newcommand{\eg}{\textit{e}.\textit{g}.}
\newcommand{\ie}{\textit{i}.\textit{e}.}

\definecolor{blue(ncs)}{rgb}{0.0, 0.53, 0.74}
\ifCLASSINFOpdf
\else
\fi

\begin{document}


%
\title{A 1Mb mixed-precision quantized encoder for image classification and patch-based compression}
%
%
%

\author{Van Thien Nguyen, William Guicquero, and Gilles Sicard 
\thanks{Van Thien Nguyen, William Guicquero, and Gilles Sicard are with the CEA-LETI, University Grenoble Alpes, F-38000 Grenoble, France (e-mail: vanthien.nguyen@cea.fr, william.guicquero@cea.fr, gilles.sicard@cea.fr).}
}

\makeatletter
\def\ps@IEEEtitlepagestyle{
  \def\@oddfoot{\mycopyrightnotice}
  \def\@evenfoot{}
}
\def\mycopyrightnotice{
  {\footnotesize
  \begin{minipage}{\textwidth}
  \centering
  Copyright~\copyright~2022 IEEE. Personal use of this material is permitted. However, permission to use this material \\ 
  for any other purposes must be obtained from the IEEE by sending an email to pubs-permissions@ieee.org. DOI: \href{https://doi.org/10.1109/TCSVT.2022.3145024}{10.1109/TCSVT.2022.3145024}
  \end{minipage}
  }
}

\maketitle

\begin{abstract}
Even if Application-Specific Integrated Circuits (ASIC) have proven to be a relevant choice for integrating inference at the edge, they are often limited in terms of applicability. In this paper, we demonstrate that an ASIC neural network accelerator dedicated to image processing can be applied to multiple tasks of different levels: image classification and compression, while requiring a very limited hardware. The key component is a reconfigurable, mixed-precision (3b/2b/1b) encoder that takes advantage of proper weight and activation quantizations combined with convolutional layer structural pruning to lower hardware-related constraints (memory and computing). We introduce an automatic adaptation of linear symmetric quantizer scaling factors to perform quantized levels equalization, aiming at stabilizing quinary and ternary weights training. In addition, a proposed layer-shared Bit-Shift Normalization significantly simplifies the implementation of the hardware-expensive Batch Normalization. For a specific configuration in which the encoder design only requires 1Mb, the classification accuracy reaches 87.5$\%$ on CIFAR-10. Besides, we also show that this quantized encoder can be used to compress image patch-by-patch while the reconstruction can performed remotely, by a dedicated full-frame decoder. This solution typically enables an end-to-end compression almost without any block artifacts, outperforming patch-based state-of-the-art techniques employing a patch-constant bitrate.

\end{abstract}

\begin{IEEEkeywords}
hardware-algorithm co-design, quantization, pruning, autoencoder, patch-based image compression 
\end{IEEEkeywords}

%

\section{Introduction and context}
\IEEEPARstart{T}WO main branches of research are pushing forward Deep Neural Networks (DNNs) into smart embedded systems and mobile devices \cite{sze_efficient_2017}. Indeed, we can state that there could be antagonist objectives depending on the targeted applications (level of versatility and programmability), the available power consumption of the system, the complexity of the inference task and its required quality of service (\eg\, level of accuracy). 

The first branch aims at developing generic hardware platforms dedicated to multi-purpose DNNs with the highest TOPS/W as the main Figure of Merit. Note that, lots of works have been done in that direction. For instance, we can non-exhaustively mention the possible simplification of the expensive matrix-vector multiplication task, using either FFT-based computing \cite{mathieu_fast_2014}, Strassen \cite{cong_minimizing_2014} or Winograd \cite{lavin_fast_2016} algorithms. Recent works improve the performance and energy efficiency of accelerators by focusing on three main key points: reconfigurability, sparsity and weights bitwidth. Within the scope of a reconfigurable platform, we can cite \cite{tu_deep_2017}, a deep neural architecture enabling highly reconfigurable patterns while targeting a wide range of models with a limited power budget of 479mW. Some other works also introduce specific architectures that limit data movement in CNN. \cite{chen_eyeriss_2016} proposed Eyeriss - a spatial architecture with row-stationary dataflow that takes advantage of local data reuse mechanism even exhibiting a lower overall chip power consumption of 278mW. On the other hand, \cite{aimar_nullhop_2019} handles sparsity with binary maps and arrays for nonzero values, skipping null activations to reach a power consumption of only 155mW. Variable bitwidth computations allow good trade-offs between accuracy and memory/power budgets as depicted in \cite{lee_unpu_2019} that presents a DNN accelerator supporting variable weights bitwidth precision from 1 to 16 bits, with a power consumption ranging from 3.2mW to 297mW (depending on the master clock frequency and power supply). \cite{yue_152_2021} proposed a computing-in-memory neural network processor efficiently dealing with sparsity and weight storage on SRAM. \cite{kim_resource-efficient_2021} presents a resource-efficient architecture for BNN inference accelerator, processing blocks in an output-oriented manner while skipping redundant operations. 

The second branch is more related to ASICs (Application-Specific Integrated Circuits) that are dedicated to only certain tasks, considering both the hardware specifications and the algorithm perspectives proposing co-optimized designs. This promotes compact network topology design with improved power consumption efficiency while capping algorithmic loss of accuracy. For instance, \cite{andri_yodann_2016} proposed an accelerator optimized for binary-weights CNN with a power consumption of 895\si{\micro W}. Using quantization technique in both analog and digital domain, Kim et al. \cite{kim_ultra-low-power_2019} proposed an always-on face recognition processor integrated with a CMOS image sensor (CIS) consuming about 620\si{\micro J} per inference. \cite{verdant_30w5fps_2020} even proposed a sub-10\si{\micro W} QQVGA imager enabling both motion detection with background estimation and face recognition using XOR-based edge extraction combined with a SVM. \cite{lefebvre_77_2021} proposed a QQVGA imager which can operate on convolution mode with ternary-weighted filters and Haar-like detection mode, under less than 206\si{\micro W}. \cite{8911327} presented a CNN-based accelerator that can serve for both face detection and facial landmarks localization. More recently, \cite{knag_617_2020} presented a digital Binary Neural Network (BNN) chip achieving a power consumption of 5.6mW.  Low-precision CNN processors were also applied to real-time object detection task in \cite{8543599} and \cite{9181590}. In particular, \cite{9181590} adopted a mixed data flow for each layer along with an intra-layer mixed weights precision quantization, in which the weights were decomposed to a dense binary kernel and a sparse 8-bit kernel in order to improve the accuracy/compression ratio trade-offs.

Please refer to \cite{Guo_NNAcc} for a more exhaustive review of published works on DNN accelerators (reporting their relative energy efficiency in terms of TOPS/W so as the types of quantization used and available operands).
\\

\textbf{Motivations:}
\\
Taking into consideration ASIC design issues, the work presented in this paper has a dual purpose: 
\begin{itemize}
\item increase the application-versatility of an Image and Signal Processor, dedicated to classification and compression;
\item improve the hardware efficiency and algorithmic performances trade-off (\ie\, inference/reconstruction accuracy).
\end{itemize}

\begin{figure}[htbp]
\centerline{\includegraphics[scale=0.25]{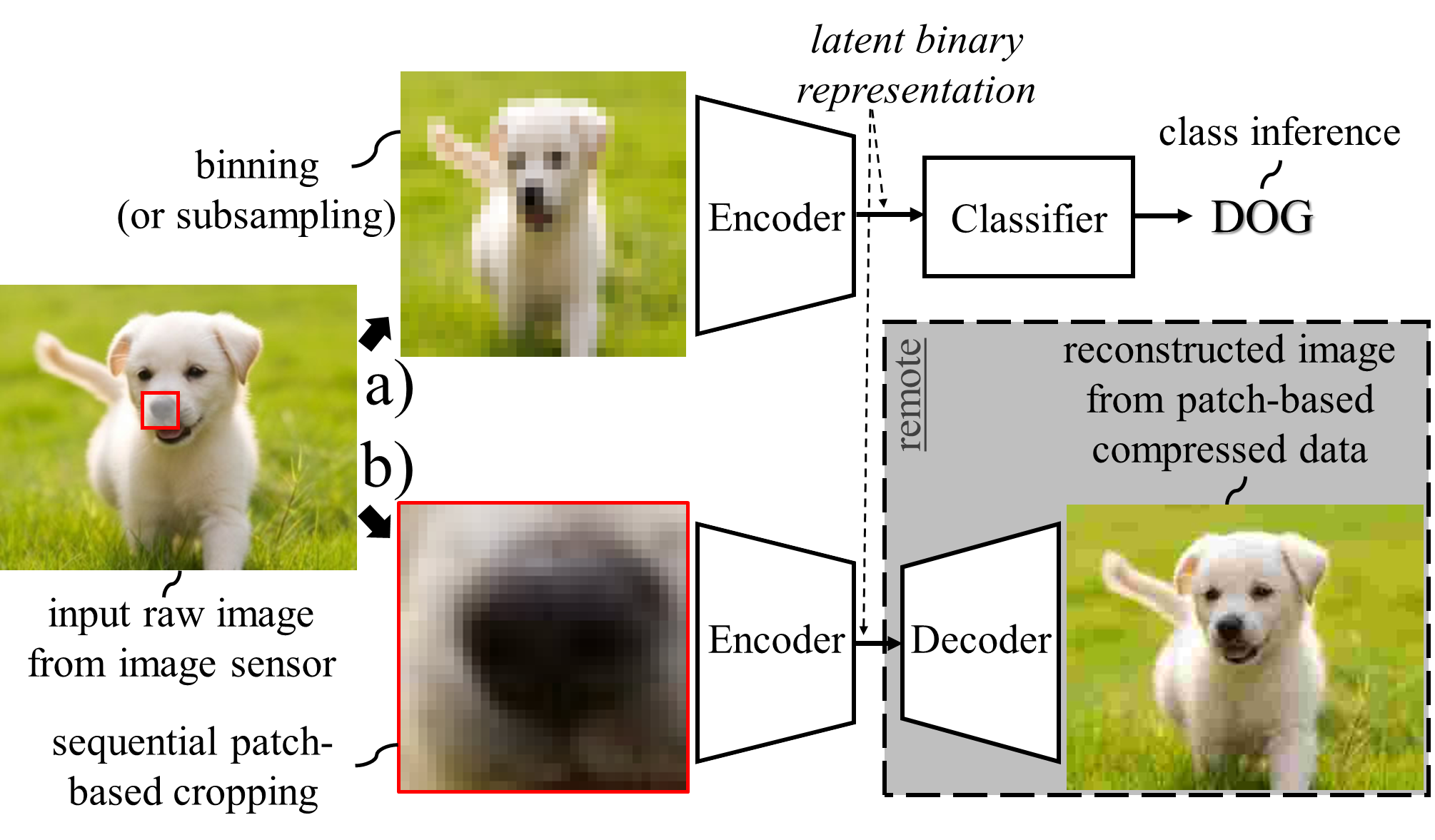}}
\caption{The joint framework for both image classification and image embedded compression and remote decompression.} 
\label{story}
\end{figure}

\begin{figure}[htbp]
\centerline{\includegraphics[scale=0.25]{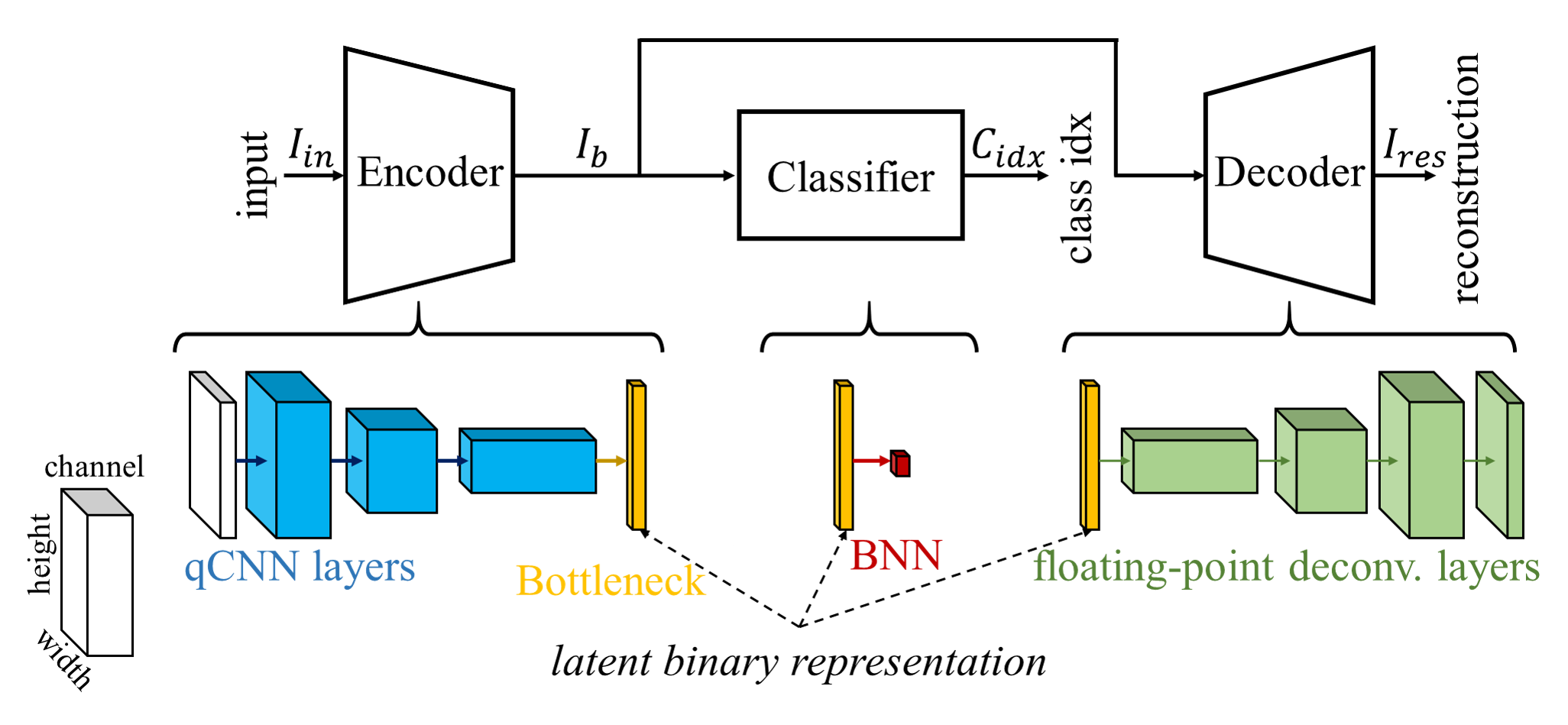}}
\caption{Schematic description of our framework involving neural network topology parts.} 
\label{story2}
\end{figure}

To this end, we propose a hardware-compliant mixed-precision encoder and its decoder counterpart. The main advantage of the proposed encoder topology is that it can be declined for both HW-light embedded inference (Figure~\ref{story}, a) path) and image compression tasks (Figure~\ref{story}, b) path) with proper weights trained separately for each application. Indeed, for each task, the mixed-precision quantized encoder will load the corresponding weights (thanks to its \textbf{reconfigurability}) to process and output discriminant patterns as a latent binary representation of the data. From this binary coding, we can thus either use directly a classifier to perform the embedded inference or a remote decoder network (PURENET) for image recovery (see Figure~\ref{story2}). \\

\textbf{Contributions:}
\begin{enumerate}
\item We introduce a mixed-precision encoder design with reconfigurability that may serve for both image compression and classification. It is noteworthy to mention here our novel adaptive quantization framework in which the quantizer is tuned according to the data during the training procedure such that the histogram of quantized levels is equally partitioned, namely histogram-equidistributed quantization. This framework is then applied it to the quinary and ternary weights in the encoder. Besides, the Batch Normalization obstacle is successfully replaced by a layer-shared Bit-Shift operation. We also propose a Half-wave Most-Significant-Bit (HWMSB) function for 2-bit activation with favorable hardware compatibility.  
\item We propose a novel decoder called PURENET that takes as input the patch-based binary measurements from the quantized encoder. To our knowledge, this work presents one of the first low-precision quantized encoder for patch-based image compression. Our experiments show that image decompression can be performed without block artifacts at low bitrate.  
\end{enumerate}

The rest of this paper is divided as follows: Part II) presents related works, Part III) details our nonlinear quantized encoder design, including details of proposed algorithmic enablers. Part IV) describes the network topology of the Decoder for image reconstruction. Part V) presents simulation results on both image classification and image compression tasks. Two additional appendices are here to support our claims, reporting benchmarks regarding the effect of the encoder bottleneck and weight-related memory versus inference accuracy.

\section{Related works}

 Weights and activations quantization, connectivity pruning and alternatives to the Fully Connected layer bottleneck are usually used as common techniques to facilitate the implementation of an Artificial Neural Network in terms of hardware mapping. Apart from algorithmic enablers for inference, the last part of this section depicts previous works on  image reconstruction from patch-based compression. 
\subsection{Weights and activations quantization}
Networks quantization compresses the model by reducing the bitwidth of weights and/or activations. The mapping from floating-point values to fixed values may be linear or logarithmic\cite{miyashita_logquant}. In Moons et al. \cite{moons_minimum_2017}, a generalized Q-bits quantization is proposed to study energy-accuracy trade-offs. In the most extreme case, networks can be trained with binary weights and activations as proposed in BNN \cite{hubara_binarized_2016}. For instance, $\{ +1, -1\}$ matrix-to-matrix multiplications convert the power-costly MAC operations to the hardware-friendly XNOR and bitcount operations. XNOR-Net \cite{rastegari_xnor-net_2016} namely gives rise to a scaling factor besides binary values to approximate activations/weights. In a more relaxing framework, Ternary Weights Networks (TWN)\cite{li_ternary_2016}, \cite{zhu_trained_2017} constraint the weights to $\{-1, 0, +1\}$ values. Although increasing 1 bit ($\approx 0.6$ bit in terms of entropy) for representing weights compared to the binary case, they allow zero values that attenuate the impacts of non discriminant neuronal activations.\cite{uhlich_mixed_2020} shows that mixed precision DNNs with learnable parametrized quantizers outperform DNNs with fixed homogeneous quantizers. However, such a learnable quantizer will not be in the scope of this work due to its hardware-related requirements. The common point of all the aforementioned techniques is that the input of the first layer and the output of the last layer are not quantized, as they play a key role in network's performance.

\textbf{Quantized step in symmetric linear quantization:} Although there exists several works on proposing non-linear quantization schemes, the hardware feasibility of those quantizers is still questionable. The linear symmetric quantization, on the contrary, may be easily adapted to an inference at the edge. However, the uniform step in this case is a crucial parameter that may jeopardize the model's performance if it is not properly chosen. Ternary Weight Networks \cite{li_ternary_2016} choose a step of $\Delta = 0.7\frac{\sum_{i=1}^{n} |W_i|}{n}$ where the mean-absolute norm factor of $0.7$ is chosen assuming the weights distribution is uniform or gaussian. Learning the quantizer is also an active approach for this subject, for example the recent works of \cite{louizos_relaxed_2019},\cite{zhao_linear_2020} and \cite{Esser2020LearnedSS}. In a different way from these works, in this paper, we propose to automatically adapt the quantized step based on layer-wise weights histogram equalization, in order to approximately maximize the entropy of quantized weights.     

\textbf{Batch Normalization in Quantized Neural Networks (QNNs):} An important remark about QNNs is that they generally fail to properly converge without Batch Normalization (BN) \cite{ioffe_batch_2015}. However this property becomes an obstacle for efficiently deploying QNNs in embedded systems, because the affine transform of BN at inference stage is still an expensive operation from the hardware point of view. Due to the crucial role of BN during the training process, there is no surprise that previous works either kept BN in full precision or avoided to use it. \cite{bankman_always-38_2019} proposed to fuse the batch normalization parameters into the kernel and bias of BNNs and perform an Addition with the 9b equivalent bias. Riptide\cite{Riptide} even approximated the variance of the BN and the scale terms of XNOR-Nets by bitshift operations, and combined them to embed both the magnitudes of weights and normalizations. Another scheme is Sign Comparison which absorbs the BN into the Sign function by an equivalent comparison between the pre-activation and the bias obtained after training\cite{valavi_64-tile_2019},\cite{jia_bitscale_imc}. However, both of these two approaches still kept the presence of the bias which surely introduced an additional module in terms of hardware implementation. Besides, these specific schemes can only be applied in the case of activation binarization, otherwise in the case where BN is followed by a multi-bit quantization, they are no more applicable.


In our work, we also have a mixed-precision approach in order to improve the harware/algorithm trade-offs. Unlike \cite{9181590} which applied an intra-layer mixed precision, we apply a layer-homogeneous quantization where we dedicate multi-bit quantization for weights and activations of the very first layers. The quantization process for activations should enable a hardware-friendly implementation mapping. Additionally, BN alternatives should not degrade network's performance so much, \eg\, smaller than 1\%  compared to standard BN. Based on these requirements, three elements will be introduced: a generic framework for multi-bit quantization based on histogram which is applied to the case of quinary and ternary weights in our encoder; a Most-Significant-Bit (MSB) function for activations quantization that extracts the relative position of the most significant bit; and finally a BitShift-based Normalization (BSN) that approximates the affine transform of the whole BN layer at testing time by a single power-of-2 rescaling, without the introduction of additional biases. 
\subsection{Inter-layer connectivity pruning}
Network pruning aims at reducing the number of operations, operands and weights in models along with overfitting, based on the assumption that fully connected layers --once trained-- exhibit sparsity and redundancy. Han et al.\cite{han_learning_2015} proposed a three-step weights pruning strategy including connection training, threshold-based weight pruning and retraining. However, a magnitude-based pruning strategy may result in irregular structures increasing hardware complexity to be properly taken into account. Therefore, several works focus on structural pruning with regularizer during training. Examples include channel-pruning\cite{he_channel_2017}, structural-aware pruning\cite{wen_learning_2016} and  budget-aware training \cite{lemaire_structured_2019}. In particular, \cite{8693518} introduces a novel hardware-compliant pruning framework where the weights are pruned within the weights-fetching groups. Besides the sparsity learning approach,\cite{mittal_recovering_2018} shows that a pre-defined random pruning can preserve the accuracy of its unpruned counterpart due to the DNN plasticity.  

In addition to conventional pruning methods mentioned above, we can also cite some works aiming to reduce the cost of standard convolutions. MobileNet \cite{howard_mobilenets_2017} introduced depthwise separable convolution, \ie\,  a single filter per input channel, followed by a regular $1\times1$ conv layer. ShuffleNet\cite{zhang_shufflenet_2018} simplified the fully-connected pointwise convolution in a Group-wise manner and a channel shuffle to equidistribute information and enhance correlation between channels. 
 
In our work, we will similarly adopt a pre-defined structural pruning technique to reduce complexity of convolution layers (memory and computations) in the context of a quantized CNN. While ShuffleNet applies Group Convolution to 1$\times$1 kernels, we propose to directly perform Group Convolution to 3$\times$3 kernels on top of the convolutional layers. 
\subsection{Alternatives to the Fully Connected layer (CNN bottleneck)}
Traditional deep CNN architectures consist of a convolutional block followed by one or many Fully Connected (FC) layers to perform the classification. Although it has shown the success on various datasets, these FC layers, in particular the very first FC layer (CNN bottleneck), often hold a huge percentage of model's parameters. For resource-constrained applications, this becomes a main limitation for the CNN deployment. Besides, these FC layers are also a factor that is prone to overfitting. There exists several methods replacing the first FC layer to achieve a better efficiency. Global average pooling is first introduced in\cite{Lin2014NetworkIN} that outputs only one value per feature map. In the same perspective, \cite{ChannelNetsCA} proposed a 3D convolution in which only local feature maps are combined to get the prediction for each class. Both these two approaches dramatically reduces the number of parameters while achieving a competitive performance compared to FC-based classifier. They all demonstrate that the FC layers exhibit redundancy and can be replaced with an acceptable loss of accuracy and generalization of the model. Indeed, both global average pooling and 3D convolutions are just a formulation of linear transformation with reduced support size compared to a dense support of FC. However, while the difference between these classifiers and FC-based classifiers is not significant in the context of full-precision models, there may be a clear gap in the specific case of quantized neural networks.

Another direction to reduce the number of parameters of FC layers is to use a fixed transform whose weight parameters are predefined. The deployment of such a model with fixed parameters is more suitable for devices with limited resources. \cite{FixYC} shows that any fixed orthogonal matrix can be used to efficiently replace a learnable FC. Even if it does not reduce the total number of operations, they demonstrate that a Hadamard matrix can improve the efficiency in terms of hardware implementation. Besides, the use of pseudo-random deterministic projections can also be a good alternative because they preserve the linear separability while being hardware-implementable, this without additional memory needs \cite{benjilali_hardware-friendly_2019}.

In this work, we propose an alternative to the first FC layer with input data of shape $H\times W \times C$ and output of $C$ hidden units. Our idea basically consists in using a depthwise convolution (DWConv) of kernel size $H\times W$ followed by an FC layer without activation in-between. Later experiments show that this approach achieves highly competitive results. 

\subsection{Autoencoder for patch-based image compression}
Patch-based (or block-based) encoding is preferred to its full-resolution counterpart, as it reduces the processing complexity and storage-related costs. Besides standards such as JPEG \cite{JPEG}, there exists a lot of other patch-based compression schemes, from dimensionality reduction using block-based Compressed Sensing (BCS) \cite{BCS}, \cite{oike_cmos_2013} to deep learning-based approaches \cite{Wu2020LearnedBH}. In the BCS framework, the high-dimensional image is divided into non-overlapping patches which are then projected separately onto a low-dimensional space via a common projection matrix, therefore we can consider it as a linear encoding scheme. The reconstruction of the entire image from these patch-based measurements can be done using iterative algorithms \cite{beck_fast_2009}, \cite{guicquero_high-order_2015} or recent learning-based methods \cite{kulkarni_reconnet_2016}, \cite{yao_dr2_2019}. On the other hand, neural network-based algorithms have been used as an alternative approach for image compression. Toderici \etal \, firstly introduced a Long Short-Term Memory (LSTM)-based autoencoder \cite{Toderici_16} enabling a patch-based $32\times32$ thumbnails compression at variable compression rates, which encodes the residual error between the current reconstruction and the original image at each iteration. This framework was then developed and applied to full-resolution image compression \cite{toderici_full_2017} with the intensive use of an LSTM-based entropy coder to capture long-term dependencies of patches. Having the same perspective, \cite{Baig2017LearningTI} introduced a progressive encoding scheme exploiting dependencies between adjacent patches. Although achieving favorable results at shallow compression rates, these Recurrent Neural Network (RNN)-based autoencoders are not suitable for being deployed in resource-constrained systems due to their complexity and hardware incompatibility. Also, these methods are not purely patch-based as they make use of spatial coherence between adjacent regions to --more efficiently-- encode every patch and alleviate the block artifacts. \cite{Li_importance_map} and \cite{8531758} proposed deep learning image compression techniques while adapting the region-based bitrate accordingly to the local content of image. Note that even if such an adaptive bit allocation would improve the overall performances, it does not fit within our hardware-restricted and patch-by-patch compression framework. Ensemble learning is recently deployed in \cite{9109567} which uses several networks of the same structure but different parameters, and a network is selected when encoding each image block before signaling to the decoder. 

\textbf{QNNs for image compression:} Quantizing the weights and activations has been showing excellent results on semantic tasks such as image classification and object detection. However, representing the weights and activations using a low precision causes a considerable loss of information at pixel level, which prevents QNNs from achieving good performances for image compression or super resolution. This explains the absence of quantized models for these pixel-level tasks in the related state-of-the-art. We can only cite some works of \cite{Ma2019EfficientSR} and \cite{xin_BNN_SISR_2020} that are dedicated to super resolution. In the case of image compression, the role of pixel-level information is more crucial in encoding image data, hence there exists hardly no prior works on quantized models for this task. However, in this paper, we use the same quantized encoder topology for patch-based image compression as the one designed for image classification. We show that even with such an encoder, the reconstruction is still guaranteed by the proposed PURENET decoder with almost no block artifacts at the equivalent bitrate of 0.25 bits-per-pixel (bpp).

\section{Nonlinear quantized encoder}
Our proposed framework first targets inference for always-on embedded systems. To ease fair comparisons with state-of-the-art approaches, we use the basic but reproducible CIFAR-10 image classification dataset \cite{krizhevsky_learning_nodate}. Even though a practical application may differ for final sizing of the topology, CIFAR-10 is generally deployed to benchmark both algorithm and hardware designs \cite{bankman_always-38_2019}, \cite{valavi_64-tile_2019}, \cite{kim_resource-efficient_2021}, \cite{jia_bitscale_imc}, \cite{cai_lowbit_rram}. Based on literature reviews related to network quantization and the VGG model \cite{simonyan_very_2015}, we took the VGG-7 model for the sake of simplicity as the pivotal element for constructing our topology variants. Note that all the reported contributions here are compatible with other kinds of neural networks architectures such as  ResNet\cite{He2016DeepRL}, MobileNet \cite{Howard2017MobileNetsEC}, ...

\begin{table*}[htbp]
\renewcommand{\arraystretch}{1.1}
	\caption{NQE topology with details on layer inputs and kernels precision.}
	\begin{center}
		\begin{tabular}{|c|c|c|c|c|c|}
			\hline
			\multicolumn{2}{|c|}{Layer} & Input shape &  Weight shape & Input precision & Weight precision \\
			\hline
			\multicolumn{2}{|c|}{Conv layer}&   $32\times 32 \times 3$ &  $3\times 3 \times 3 \times F$  & 8 & 3 \\
			\hline
			\multicolumn{2}{|c|}{Conv layer 2} &  $32\times 32 \times F$ &  $3\times 3 \times F \times F$& 1 & 3 \\
			\hline
			\multicolumn{2}{|c|}{Conv layer 3} &  $16\times 16 \times F $ &  $3\times 3 \times F \times 2F$& 2 & 2 \\
			\hline
			\multicolumn{2}{|c|}{Conv layer 4} &  $16\times 16 \times 2F $ &  $3\times 3 \times 2F \times 2F$& 1 & 2 \\
			\hline
			\multicolumn{2}{|c|}{Conv layer 5} &  $8\times 8 \times 2F $ &  $3\times 3 \times 2F \times 4F$ & 2 & 1 \\
			\hline
			\multicolumn{2}{|c|}{Group Conv ($G=4$)}  & $8\times 8 \times 4F $ &  $3\times 3 \times F\times 4F$ & 1 & 1 \\
			\hline
			\multirow{2}{*}{Bottleneck}& {Depthwise Conv} &  $4\times 4 \times 4F $ &  $4\times 4 \times 1 \times 4F$ & \multirow{2}{*}{1} & \multirow{2}{*}{1} \\
			\cline{2-4}
			& FC  &  $1\times 1 \times 4F $ &  $4F\times 4F$ & &  \\
			\hline
			\multicolumn{2}{|c|}{FC} &  4F  &  $4F \times 10$ & 1 & 1 \\
			\hline
		\end{tabular}
		\label{hardware_details}
	\end{center}
\end{table*}

\begin{figure}[htbp]
	\centerline{\includegraphics[scale=0.25]{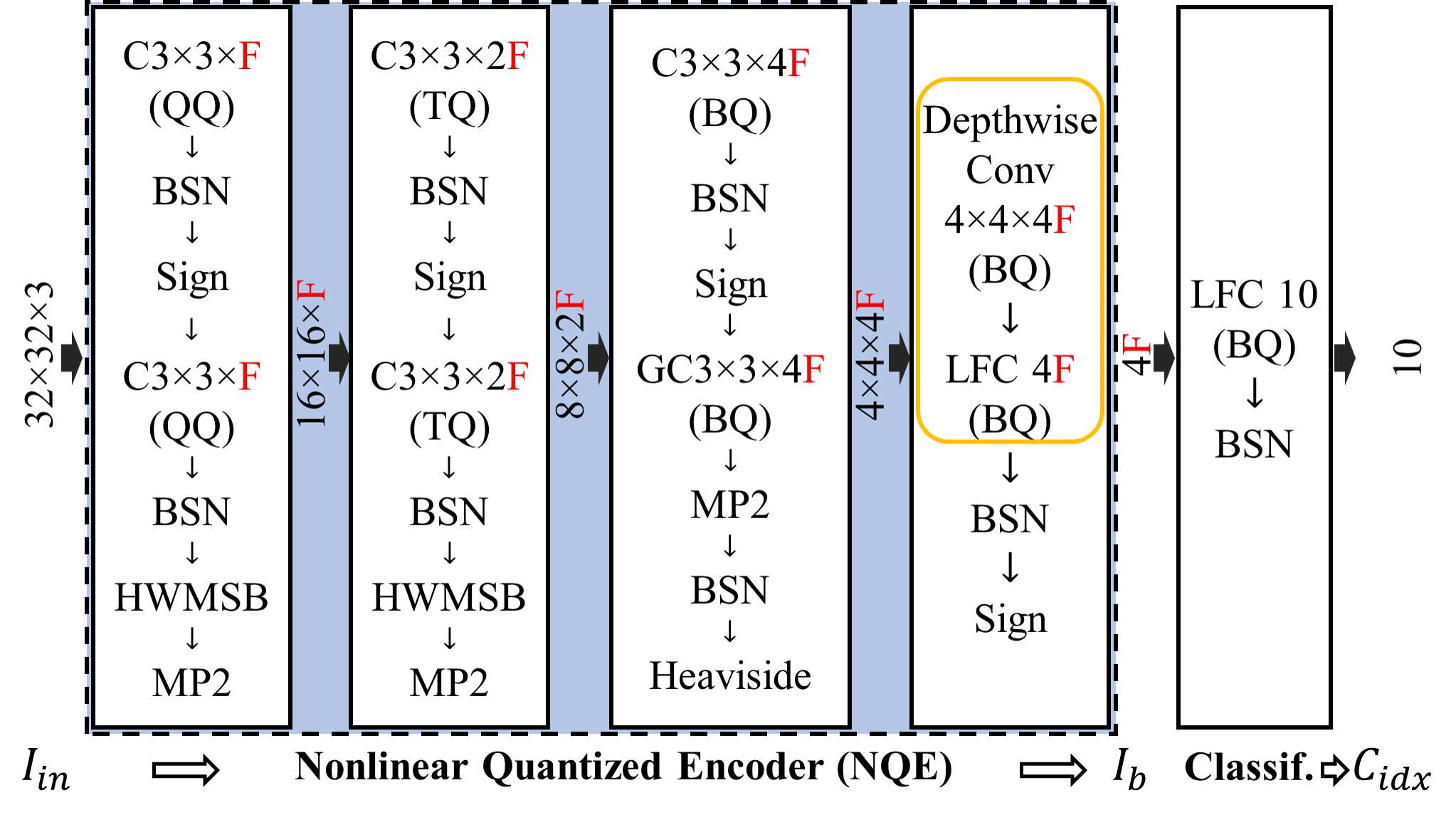}}
	\caption{Topology of the Nonlinear Quantized Encoder (NQE) + Classifier. {\color{red}F} in red stands for the hyperparameter corresponding to the size scale of the feature map (\ie\, the number of the feature map of the first convolution module). GC stands for Group-wise Convolution of 4 groups.} 
	\label{classifier}
\end{figure}

Figure~\ref{classifier} depicts our hand-crafted mixed-precision network topology integrating all proposed algorithmic enablers for the image classification task. The model is divided into 5 main modules, with 3 Convolution blocks, 1 Bottleneck layer and 1 output Classifier. In this specific setting (\ie\, the baseline model), a multi-bit quantization scheme for both activations and weights is already applied to the first two convolutional modules, \ie\, Quinary weights Quantization (QQ) for the first, Ternary weights Quantization (TQ) for the second and Binary weights Quantization (BQ) for all the rest. Since the first conv layers are crucial to extract meaningful information therefore more bits for them leads the network to better performance trade-offs. Note that the very first conv layer additionally embeds channel-wise biases to properly enable image dynamics feature extraction. We also put HWMSB activations at the end of the first two convolutional blocks and the heaviside activation at the end of the third block, as having zero values --instead of signed ones-- helps learning to discriminate better data features. The first non-convolutional layer is the bottleneck which holds the half of model's parameters in its baseline format (\ie\, a FC layer). This first FC is thus replaced by a depthwise convolution followed by a far smaller FC. These first fours blocks constitute together what is denoted a Nonlinear Quantized Encoder (NQE). The final block as for it, is a classifier composed of one single FC. From a top-level system view, the NQE will perform the image compression over non-overlapped $32\times32$ patches while in classification mode it is combined with the last-stage classifier. We also want to stress that applying $2\times2$ MaxPooling (MP2) on a binarized tensor results in a tensor with almost all ones, confusing the training procedure when choosing the argmax positions during backpropagation. However, applying MP2 over quantized values is more hardware-friendly than over full-precision values. In Figure~\ref{classifier}, MP2 are thus put after HWMSB but before Heaviside keeping in mind that from the hardware point of view (\ie only for feedforward pass) this last order can be reversed. Note that details of the layer weights and input configuration are described in Table~\ref{hardware_details}.

\subsection{Histogram bins equidistributed quantization}
A linear symmetric quantization of restricted range of odd $n$ discrete values ($n>2$) can be described as follows:
\begin{equation}
q(x; \Delta) = \frac{2}{n-1} \textrm{Clip}\left( \lfloor\frac{(n-2)x}{2\Delta}\rceil, \frac{1-n}{2}, \frac{n-1}{2} \right),
\label{linear_quant}
\end{equation}
where the parameter $\Delta$ controls the quantization step and clipping which is defined as $\textrm{Clip}(x; a,b) = \textrm{min}(\textrm{max}(x, a), b)$ with $a < b$. In this formulation, all quantized values $q \in \frac{2}{n-1}\{\frac{1-n}{2} + 1, ..., 0, ..., \frac{n-1}{2}-1\}$ are uniformly distributed inside the interval $(-\Delta, \Delta)$, bounded to $\pm 1$. Indeed, thanks to the $\frac{2}{n-1}$ scale factor, all the quantized values are shrunk in the interval $[-1, +1]$. 

Figure~\ref{linear_quant} depicts the case of $n=3$ (ternary quantization\cite{li_ternary_2016}) and $n=5$ (quinary quantization). This formulation deeply depends on the quantization step parameter $\Delta$, and how to choose an optimal value for it is still an open question. In the Ternary Weight Networks\cite{li_ternary_2016}, $\Delta = \tau \frac{\sum_{i=1}^{n} |W_i|}{n}$ where the mean-absolute norm factor $\tau = 0.7$ is estimated by assuming the weights distribution to be either Uniform or Gaussian. We argue that in many cases, depending on the layer's position, the model topology, the training procedure and the inference task, these assumptions are usually invalid, resulting in a sub-optimality and an unbalance between the number of weights at each quantized level. For example, if the floating-point values concentrate mainly around zero, it is likely that there may be too many zeros over other quantized values. 

Starting from this analysis, we propose to estimate the mean-absolute norm factor based on the histogram of the proxy weights, such that the quantized weights are distributed with nearly equi-probabilities. This way, we aim to increase the entropy of the quantized weights, and consequently, the diversity of the output values. Let us denote $q_0, q_1, ..., q_n$, $n+1$ points which divide the weights histogram into $n$ equal quantiles, where $q_0$ and $q_n$ are the minimum and maximum values, respectively. Observing that during the training procedure, the weights distribution may change, but the median value usually stays around zero, we assume that these quantiles are approximately symmetric. Concretely, to equidistribute the histogram bins of quantized weights, the parameter $\Delta$ and $\tau$ are re-estimated and updated according to the updated proxy weights after each training epoch such that these quantile points correlate with the thresholds in the quantization function. To update the proxy weights during each epoch, we keep using the STE\cite{bengio_estimating_2013} strategy with a clipped identity, while $\Delta$ and $\tau$ remain fixed between mini-batches. For instance, in the case of ternary quantization, the points $q_1$ and $q_2$ in Figure~\ref{linear_quant} must match the points $-\Delta$ and $\Delta$, therefore we have the approximation $\Delta = \frac{|q_1| + q_2}{2}$. Similarly, in the case of quinary quantization, the points $q_1, q_2, q_3, q_4$ should coincide with the points $-\Delta, -\frac{\Delta}{3}, \frac{\Delta}{3}, \Delta$, therefore we have the approximation $\Delta = \frac{3(|q_1| + |q_2| + q_3 + q_4)}{8}$. From this approximated $\Delta$, we can thus estimate an equivalent mean-absolute norm factor $\tau$ (cf. \cite{li_ternary_2016}) using the following equation:

\begin{equation}
\tau = \frac{n\Delta}{\sum_{i=1}^{n} |W_i|}.
\label{norm_factor}
\end{equation}
\begin{figure}[htbp]
	\centerline{\includegraphics[scale=0.25]{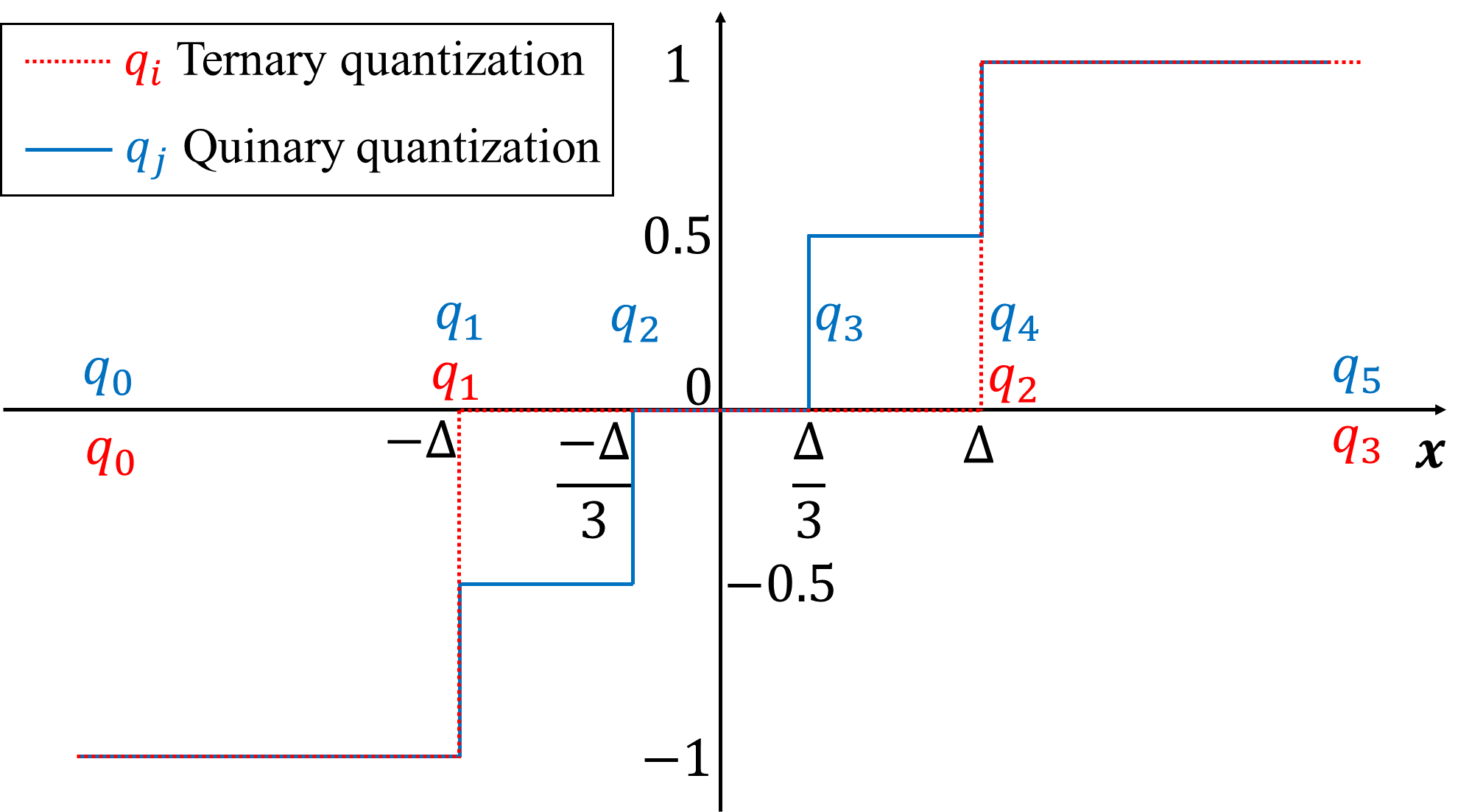}}
	\caption{Symmetric linear quantization with histogram bin equidistribution. The ternary quantization is represented with red dotted curve and red quantiles ${\color{red}q_i}$. The quinary quantization is represented with blue solid curve and blue quantile ${\color{blue(ncs)}q_j}$. For simplicity, we choose the same $\Delta$ for both two quantizations, but in practice, they usually have a different estimated $\Delta$.} 
	\label{linear_quant}
\end{figure}    

While the ternary weight networks are widely deployed in designing efficient CNNs, the quinary quantization is still quite novel. Even having only 5 over 8 quantized values possible of a 3-bit representation, a kernel of $0, \pm 0.5, \pm 1$ greatly simplifies the matrix-matrix multiplication into logical operations, bitshifts, and accumulations. It also limits the number of bits to store the intermediate values, while increasing significantly the representability of the kernels. Therefore, we apply quinarization and ternarization to the first two convolution modules only, in order to extract more meaningful information from image features while limiting the overall impact on memory needs.

\subsection{Half-Wave Most-Significant-Bit (HWMSB) activation}
To compensate additional hardware needs when using more bits for intermediate values, we propose to simply extract the position of the most significant bit of the income value. This operation advantageously embeds two wanted features, namely $log_2$ dynamic range compression and intrinsic requantization. Its original real-valued function mapped in the $[-1, +1]$ range can be defined as:
\begin{equation}
f(x)=\begin{cases}
\textrm{sign}(x) \textrm{ min}\left(\frac{4 + \textrm{log}_2 (\left|x\right|)}{3},1\right) & \textrm{if } x \geq \frac{1}{8},\\
\frac{8x}{3} & \textrm{otherwise.}
\end{cases}
\end{equation}
The MSB quantization function and its Straight-Through-Estimated (STE\cite{bengio_estimating_2013}) gradient are then described as follow: 
\begin{equation}
q(x) = \begin{cases}
\textrm{sign}(x) \textrm{ min} \left(\frac{\lfloor 4 + \textrm{log}_2 (\left|x\right|) \rfloor}{3},1\right)  & \textrm{if }|x| \geq \frac{1}{8},\\
0          & \textrm{otherwise.}
\end{cases}
\end{equation}
\begin{equation}
\frac{\partial q (x)}{\partial x} =  \begin{cases} 
\frac{1}{3\left|x\right| \textrm{ } log2}   &  \textrm{if }  \frac{1}{8} \leq |x| \leq 1,\\
\frac{8}{3}   & \textrm{if }  |x| < \frac{1}{8}.
\end{cases}
\end{equation}

When this MSB is followed by a ReLU activation, only positive values can be passed and negative values are zeroed out. We call this combination as Half-wave Most-Significant-Bit (HWMSB) activation. Unlike the MSB function, the HWMSB activation has only 4 possible output values $\{0, 1/3, 2/3, 1\}$, therefore it needs only 2 bits to represents the output data. Another advantage of HWMSB compared to MSB is that attenuating negative values improves the learning as ReLU does in several topologies. Table~\ref{tab1} reports the value mapping between the decimal, naive binary representations and the outputs. The first significant bit is assigned to the third bit on the right of the point. We call this the reference position, which is determined by the integer bias of $4$. If the input is multiplied by a power-of-2 factor before the MSB, we can absorb this multiplication into the MSB by just shifting the reference position accordingly. Note that this quantization scheme differs from existing logarithmic quantizations such as~\cite{miyashita_logquant} on these two points, first it has the integer bias of $4$ and it also zeroes out negative values. Note that the normalization factor (here, \text{3} to keep the dynamic in the wanted range) can be assigned independently to this scheme. 

\begin{table}[htbp]
	\caption{2-bit HWMSB input-output mapping with naive binary representation (sign+magnitude)}
	\begin{center}
		\begin{tabular}{|c|c|c|c|}
			\hline
			\multicolumn{2}{|c|}{\textbf{Input}}&\multicolumn{2}{c|}{\textbf{Output}} \\
			\hline
			\textbf{\textit{Decimal}} &  \textbf{\textit{Binary}}  & \textbf{\textit{Decimal}} &  \textbf{\textit{Binary}} \\
			\hline
			$ x < 0.125 $ & x.000xx/ 1.xxxxx & 0 & 0.00\\
			\hline
			$0.125 \leq x < 0.25 $ & 0.001xx & 1/3 & 0.01\\
			\hline
			$ 0.25 \leq x < 0.5 $ & 0.01xxx & 2/3 & 0.10\\
			\hline
			$ x \geq 0.5 $  & 0.1xxxx & 3/3 & 0.11\\
			\hline
		\end{tabular}
		\label{tab1}
	\end{center}
\end{table}

\subsection{Layer-shared BitShift-based Normalization (BSN)}
BN keeps a crucial role to QNNs, especially BNNs, as these networks fail to well converge without a proper rescaling. However, BN is not tractable for Deep Learning in highly constrained embedded systems, since at inference time, it consists of one full-precision addition and multiplication per scalar, which is  a computational-demanding operation. A quantized BN from scratch is definitely not as robust as the standard BN as it is difficult to estimate the appropriate scaling factor, and unfortunately involves a significantly lower network's accuracy (\eg\, much larger than 1\%). 

\textbf{BN replacement by a single BitShift:} The straightforward option considered here is we still employ the standard BN to train the model from scratch and then simplify all BN affine transforms by a single BitShift approximation, with later retraining in order to update the weights accordingly. This two-step training procedure allows preserving its accuracy performances with a high simplification of the final hardware implementation. We thus approximate the scale constants of BN layers obtained after first training in a power-of-2 fashion, that advantageously corresponds to the bitshift operation. After the training stage, BN layer has properly estimated batch statistics $\mu$ and $\sigma^2$, respectively representing the moving mean and the moving variance. Let us recall that at the inference stage, the BN consists in processing the input $x$ to provide the output $y$ as follows:

\begin{equation}
y = \gamma \frac{x - \mu}{\sqrt{\sigma^2 + \epsilon}} + \beta  \equiv \hat{\gamma} x + \hat{\beta}
\end{equation}

where --using the same notations as in \cite{ioffe_batch_2015}-- $\hat{\gamma} = \frac{\gamma}{\sqrt{\sigma^2 + \epsilon}}$ and $\hat{\beta} = \beta -  \frac{\gamma \mu}{\sqrt{\sigma^2 + \epsilon}}$ are equivalent to a scale and an offset  (\ie\, channel-shared additional weight and bias if applied after 2D convolution, and unit-shared if applied after 1D Dense layer, therefore we have a vector of different $\hat{\gamma}$ for each BN). Concretely, in our framework, we choose the 0.9-quantile value from these scales at each layer, denoted as $\tilde{\gamma}$, to serve as the unique scale for all the BSNs, approximating it as follows:

\begin{equation}
y = 2^{\lfloor log_2|\tilde{\gamma}| \rfloor} x 
\end{equation}

Note that the experimentally chosen 0.9-quantile is considered as it is a good trade-off between the maximum value that may explode the dynamic range along with the gradient, and the minimum value that may slow down the gradient update of previous layers. After replacing all BN layers of the model by this single BSN transform, we train again the network one more time. Note that for all the BSNs, we can get rid of the bias $\hat{\beta}$ as it does not improve the inference in general, at the expense of additional computations for hardware mapping. 

\textbf{Hardware implementation of BSN:} Generally, having only one bitshift-based scale replacing the whole BN layer considerably reduces the computational complexity in all cases, regardless its following quantization. While needing only 4 bits to store the bitshift scale, all input data will share a common bitshift operation, that is much cheaper than affine transforms of different scales and biases, even if theses parameters are quantized. Moreover, in the context of our model topology where we have either HWMSB, Sign or Heaviside quantization after the BSN, this becomes more advantageous. Clearly, the single bitshift keeps the sign of data unchanged, therefore, it has no impact to the results of the later Sign or Heaviside quantization. Consequently, the BSNs followed by Sign or Heaviside quantization do not need to be explicitly implemented. Similarly, we can also get rid of the final BSN as it does not change the order of the logit predictions. In addition, the bitshift scale of BSN can be intrinsically fused into the HWMSB by just shifting the reference position of the HWMSB accordingly.

\subsection{Pre-defined pruning with Group-wise Convolution (GC)}
\begin{figure}[htbp]
	\centerline{\includegraphics[scale=0.25]{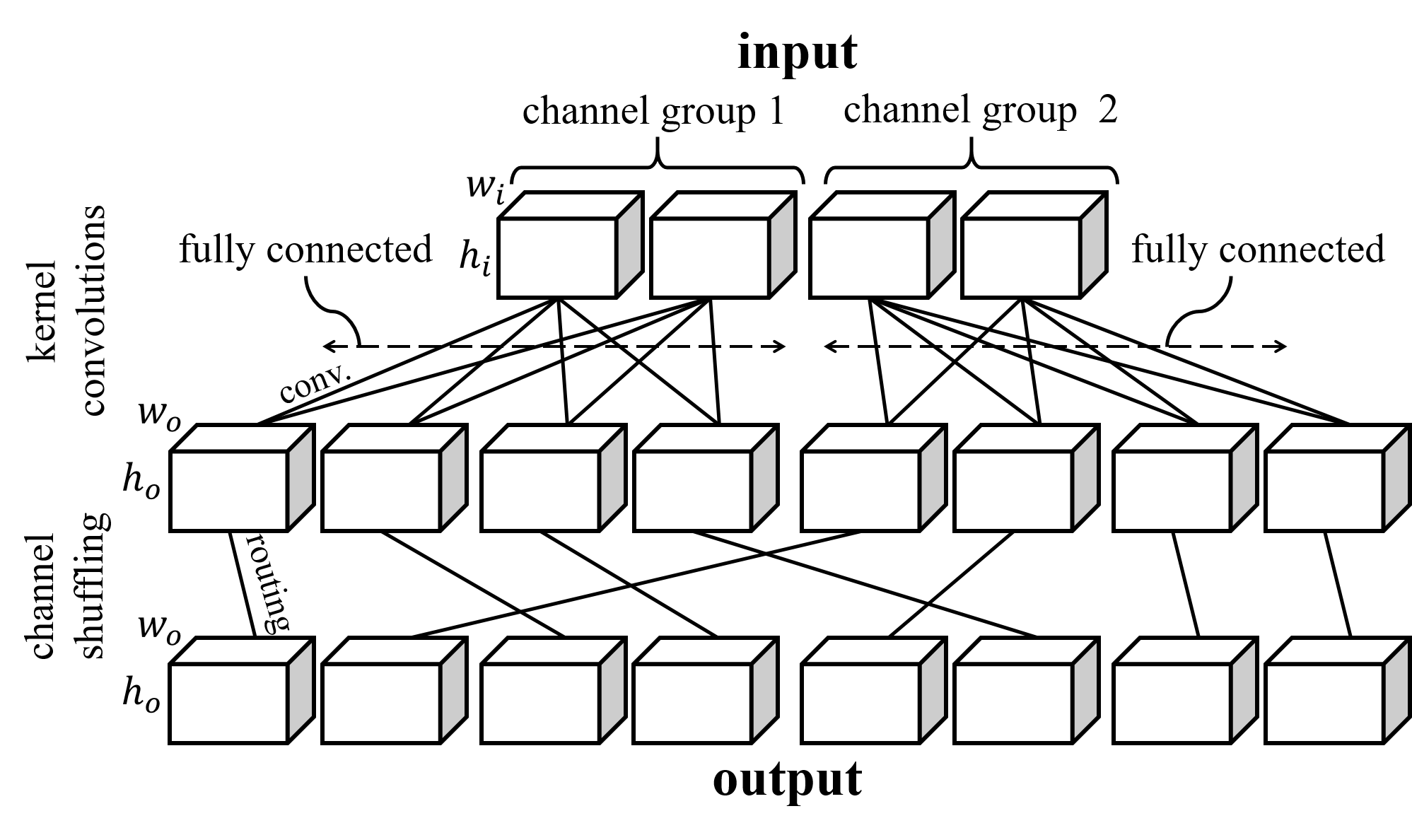}}
	\caption{Example of Group-wise convolution with a 4-channel input divided into 2 groups and 8-channel output. The intermediate values are also divided into two groups, and each convolution is performed with a kernel that takes only two input channels from the corresponding group. These output channels are then structurally shuffled.} 
	\label{gconv}
\end{figure}

Inspired by ShuffleNet \cite{zhang_shufflenet_2018} and justified by the advantages of a structurally pre-defined pruning, we propose to perform convolutions of CNN in a Group-wise manner instead of an all-channel-fully-connected conventional topology (see Figure~\ref{gconv}). This group-wise pruning is only applied to the last convolutional layer of the NQE, as it contains most of the parameters. A Group-Convolution reduces the parameters and the number of MACs by a factor equal to the number of groups. Consequently, it also reduces the memory needed for intermediate values. Compared to unstructural pruning scheme, a predefined structural pruning like the Group Convolution may be embedded directly to the hardware platform, without the need of additional memory to save the connection positions.

\subsection{Compression of the bottleneck dense layer}
In our NQE, the first affine transform performed after conv modules has an input of shape $4\times 4 \times 4F$ and outputs $4F$ hidden units. In the case of a dense layer, it will contain $256F^2$, \ie \, almost  50$\%$ of the model's parameters. Even though the weights are binary, it still requires a large percentage of the total memory needs. Therefore, replacing this bottleneck layer while preserving the model's performance is crucial for improving the overall efficiency. In this paper, we propose to firstly use a depthwise convolution of kernel size $4\times4$, to transform the 3D tensor into a vector of length $4F$. Formally, this depthwise convolution is equivalent to  a block-diagonally connected layer, with only $4\times4\times4F = 64F$ learnable parameters. Then we have a square learnable FC layer with size $4F$, therefore holding only $16F^2$ parameters. Since there are no activation between these two sub-layers, this approach is equivalent to decomposing the fully dense matrix into two sub-matrices, one for only spatial operation (depthwise convolution), the other for the combination of channels (dense layer). Consequently, the total number of parameters is $16F^2 + 64F$, which is very small compared to the initial $256F^2$ parameters of the FC version of the bottleneck. In Figure~\ref{classifier}, these two layers are surrounded by a yellow rounded rectangle, denoting that they form together an alternative to the bottleneck dense layer.  

\begin{figure*}[h]
	\centerline{\includegraphics[scale=0.5]{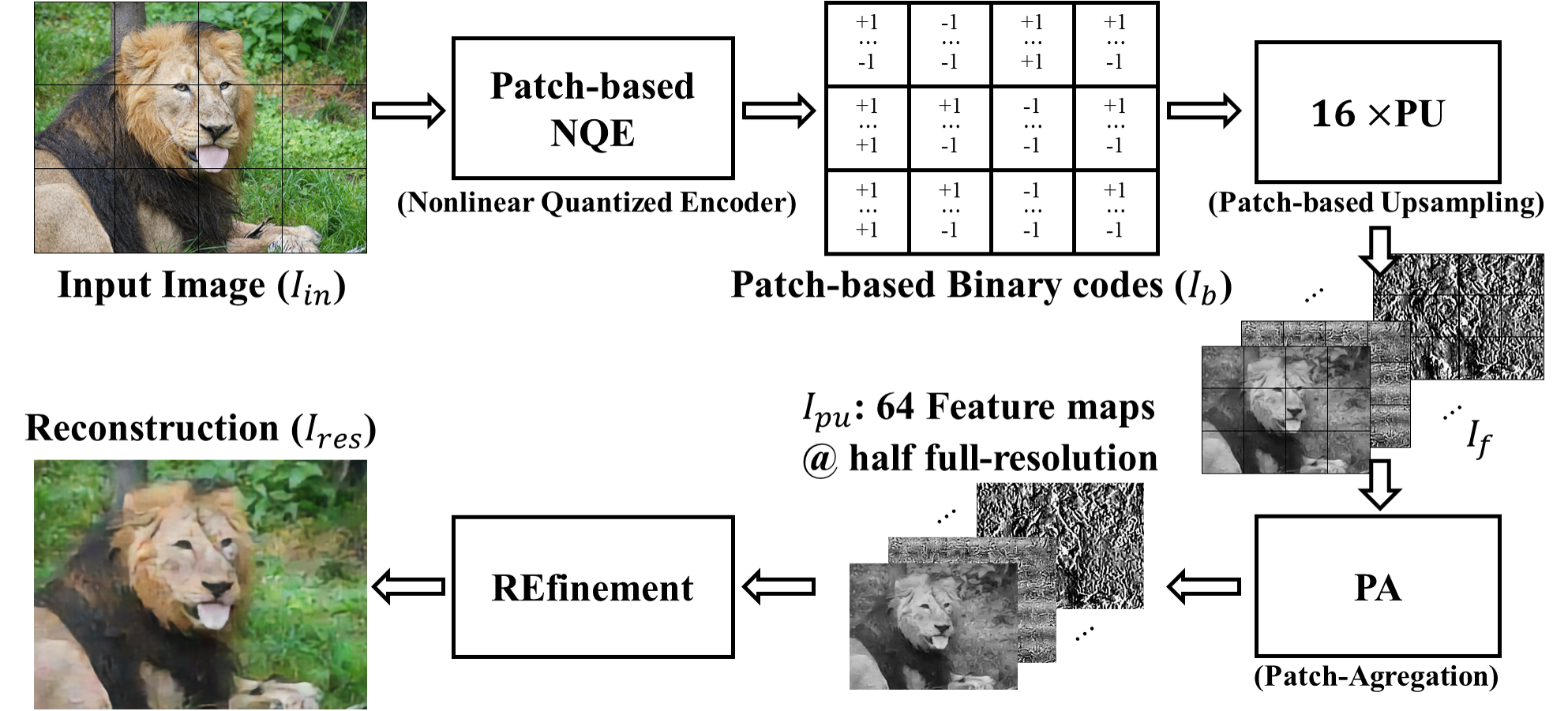}}
	\caption{The patch-based encoding with Full-Resolution PURENET decoder. Compressed binary codes are vector of length $256$, corresponding to the bitrate of $0.25$bpp. Note that without the Patches Aggregation (PA), this topology may still be applied to the decoding of patches independently. This variant without PA is denoted as patch-independent PURENET (PI-PURENET).} 
	\label{encode_decode}
\end{figure*} 

\section{Image reconstruction from patch-based quantized measurements with PURENET}
Figure~\ref{encode_decode} shows the proposed combination of an image patch-based encoder with a full-frame decoder. The large image is first divided into small patches of size $32\times 32$ and then processed by the NQE to obtain a binary representation for each patch independently. The Patch-based Upsamling (PU) module will learn to increase the spatial resolution of these codes from  $1\times1$ (vector) to $16\times16$ (\ie \, half of the final full-resolution). These patches are then aggregated together to form proxy feature maps at $\frac{1}{2} \times \frac{1}{2}$ full-resolution, which are then processed by the Refinement module to obtain the final reconstruction. The term PURENET hereafter denotes for the Patch-based Upsampling and REfinement NETwork.      

\begin{figure}[htbp]
	\centerline{\includegraphics[scale=0.25]{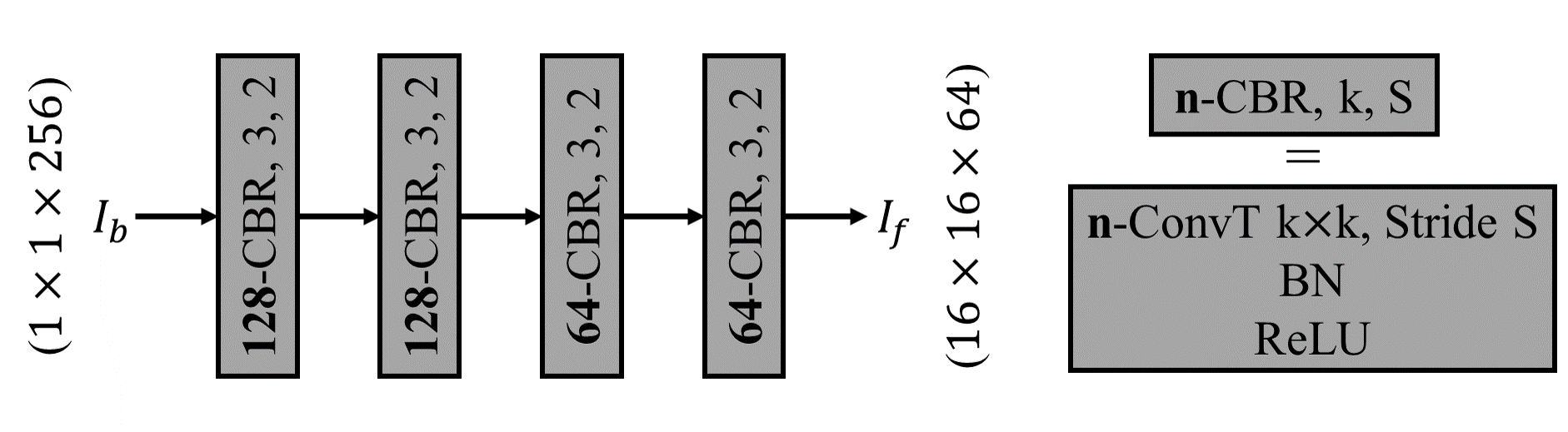}}
	\caption{PU: The Patch-based Upsampling performed using 4 CBRs (ConvT + BN + ReLU), all with a kernel size of $3\times3$ and a stride of 2.} 
	\label{upsampling}
\end{figure}

\textbf{Upsampling module:}  
In details, the Upsampling model contains 4 blocks of Transpose Convolution + BN + ReLU (CBR). Each Transpose Convolution (ConvT) is of kernel size $3\times3$ and strides of 2. Figure~\ref{upsampling} depicts the topology of this module. In PURENET, the Upsampling module is independently applied to every patch, so that the information of each patch is properly preserved and do not mix with the neighborhood. We observe that aggregating the patches before the final $2\times 2$ upsampling is an appropriate trade-off between alleviating the block artifacts and limiting the color errors due to the mixture of patches. 

\begin{figure}[htbp]
	\centerline{\includegraphics[scale=0.25]{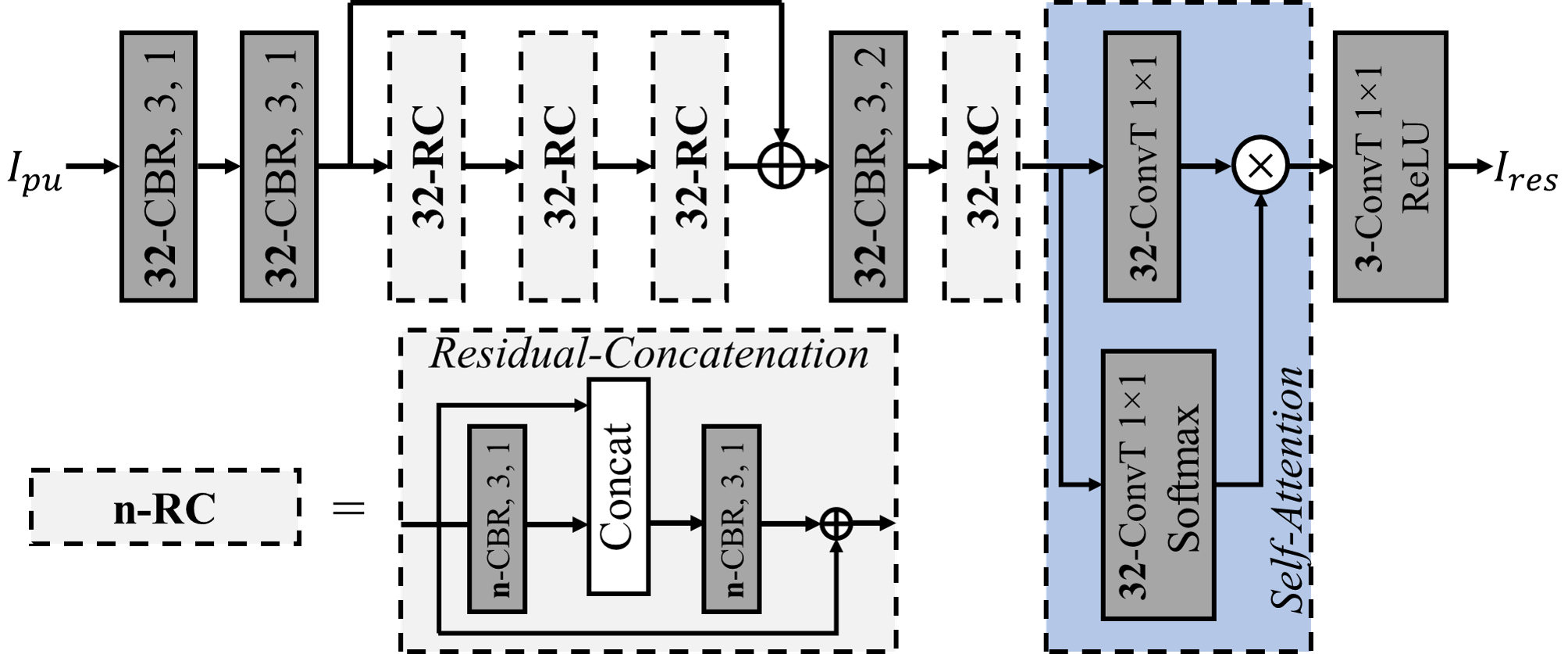}}
	\caption{RE: Refinement model architecture with Residual and Concatenation (RC) Block. The parameter $n=32$ denotes the number of feature maps.} 
	\label{refinement}
\end{figure}

\textbf{Refinement module:}
After aggregating all patches, we obtain a tensor of half resolution of the original image. The Refinement stage (Figure~\ref{refinement}) then allows smoothing the image, hence removing unwanted block-artifacts. Each patch is now reconstructed by using not only its own information but also its neighbors. In particular, this Refinement model makes use of several Residual-Concatenation (RC) blocks before the final upsampling in order to reach the original resolution. Each RC block consists of 2 CBRs, with feature maps fusion inserted in-between that concatenates the RC input with the output of the first CBR and an add-skip connection at the output. Once the last upsampling is performed, the image feature maps pass through a last RC block followed by a self-attention like mechanism. In this sub-module, each of the two branches contains a CBR with pointwise ($1\times1$) convolution, and one branch has a softmax activation at the end to normalize the response of the pixels across the channel dimension. The two branches are then combined together by an element-wise multiplication, before outputting an RGB image for the final reconstruction. To demonstrate the impact of the Refinement module, we denote also the Block-Based Decoder (BBD) which contains the PU and an additional CBR with a stride of 2 for a neighborhood-independent reconstruction of patches.

\textbf{PURENET training:} We adapt a two-stage training procedure for PURENET. Firstly, we train the NQE with the patch-independent PURENET (\ie \, PURENET without the Patches Aggregation, PI-PURENET) so that the NQE learns to compress $32\times32$ patches. After this stage, we obtain patch-based binary codes for each image which are then used as the input of PURENET in the second training stage. Finally, the pre-trained PI-PURENET is used to initialize the weights of corresponding modules (Patch-based Upsampling and Refinement) in the full-resolution mode (PURENET). 

\section{Simulation results}
In this section, we target a configuration such that the memory budget for embedded classification is only approximately of $1$Mb (with naive weights encoding), by choosing ${\color{red}F}=64$. Note that if we encode the quinary ($2.32$b) and ternary ($1.58$b) weights properly with an entropy coder, the on-chip memory may even be reduced to a sub-$1$Mb budget (which is not dealt in this paper). All the software implementations of this paper have been done in Python using a Tensorflow2 backend. 

\subsection{Image classification on CIFAR-10} 
\label{classif_report}
In order to train our models, we apply the same data augmentation scheme as proposed in \cite{lee_deeply-supervised_2015}, \ie\, a 4-sided 4-pixel padding followed by a random crop with random horizontal flip in order to provide 32$\times$32 images. For classification task, the models are trained with a batch size of 50 using a squared hinge loss and the Adam optimizer \cite{kingma_adam_2017}. 100 epochs are performed for the first training with standard BN and then 120 epochs for the fine-tuning stage, with BSN. The learning rate is initialized at $10^{-3}$ and decreased with an exponential decay with the rate of $0.8$. The reported average accuracies (Table~\ref{comparison}) are over 3 realizations for each point. 

\begin{figure}[h]
     \centering
         \begin{subfigure}[b]{0.24\textwidth}
        \includegraphics[width=\textwidth]{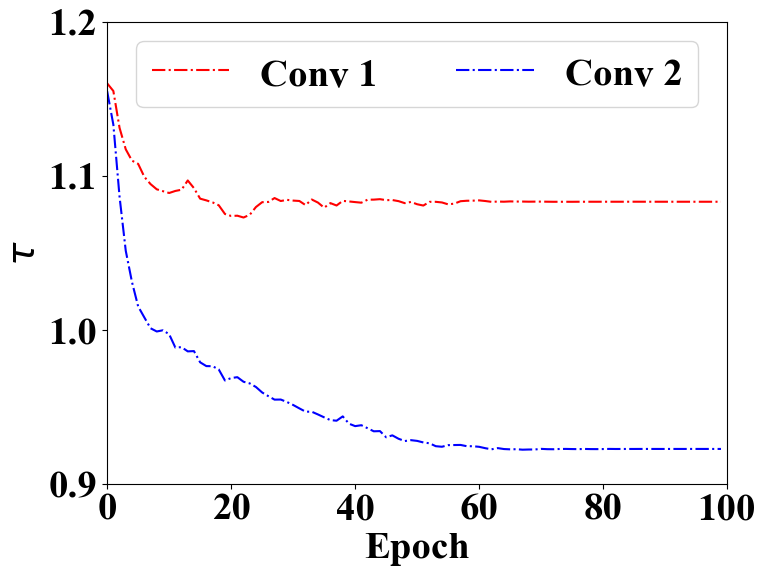}
        \caption{Quinary quantization.}
        \label{ternary}
    \end{subfigure}
    \hspace{-2mm}
    \begin{subfigure}[b]{0.24\textwidth}
        \includegraphics[width=\textwidth]{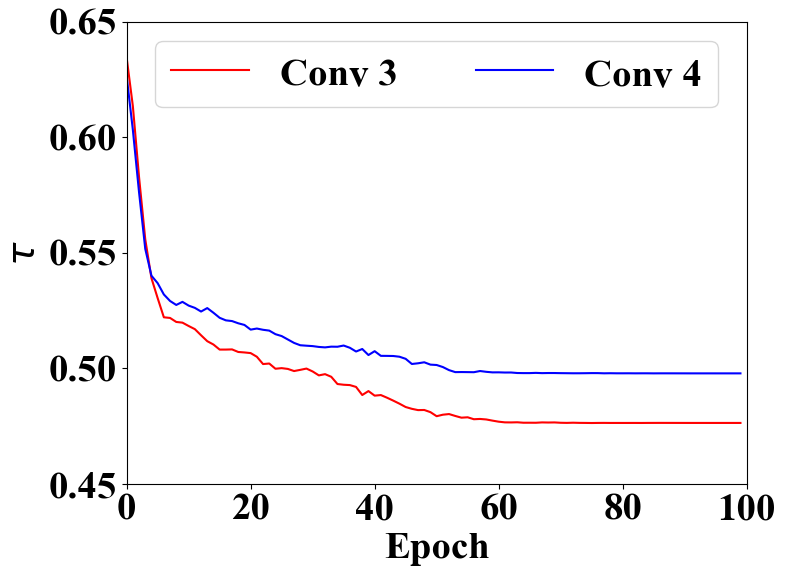}
        \caption{Ternary quantization.}
        \label{quinary}
    \end{subfigure}
     \caption{Evolution of $\tau$ during training.}
     \label{stepsize_curves}
\end{figure}

\textbf{Curve of $\tau$ during training:} Figures~\ref{stepsize_curves} shows the variation of the mean-absolute norm factor $\tau$ (defined in Equation~\ref{norm_factor}) in the case of quinary (\nth{1} and \nth{2} Conv layer) and ternary quantization (\nth{3} and \nth{4} Conv layer). As can be seen on those figures, the mean-absolute norm factor decreases before finally converging when the learning rate becomes smaller and the model itself reaches a more stable state. The final values of the mean-absolute norm factors differ from each other, even for the same type of quantization. It is worth mentioning
that in the case of second conv module (ternarization), the obtained factors ($0.476$, $0.498$) appear to be much smaller than the value reported in \cite{li_ternary_2016} (\ie\, $0.7$). 

\textbf{Comparison with prior works:} Table~\ref{comparison} provides a comparison with state-of-the-art CNN accelerators that have been recently demonstrated including both BNN and mixed-precision based designs. To measure the computational complexity, we make use of two bitwidth-aware metrics called MAC$\times$bit \cite{9401657} and Bit-Operations (BOPs \cite{WangLB20}). For a relative complex dataset like CIFAR-10, a multi-bit quantization accelerator is proved to be more robust than a binarized accelerator. An important question has thus arisen: how to design such a mixed-precision architecture with higher accuracy while lowering memory needs, computational complexity and simplifying hardware implementation? While the two mixed-precision designs in \cite{jia_bitscale_imc} and \cite{cai_lowbit_rram} obtain an accuracy larger than 90$\%$, they also contain large on-chip memory of nearly $30$ and $19$Mb along with sub-1G MAC$\times$bit and BOPs, which generally overpass the capacity of resource-constrained ASICs. On the other hand, BNN-based designs obtain lower accuracy at lower on-chip memory. For instance, the binary NQE (sharing the same model architecture but with all binary weights/activations) obtains only 82.40$\%$ accuracy with a tiny budget of both memory and computation, while our mixed-precision NQE topology achieves $87.48\%$ accuracy in average, while requiring only $1$Mb of weight parameters as well as under-0.3G MAC$\times$bit and BOPs. This improvement of 5$\%$ demonstrates the significant contribution of the mixed-precision topology to the overall performance. Compared to the reported design with the nearest accuracy\cite{bankman_always-38_2019}, our design requires $2.4\times$ less on-chip memory and $3.5\times$ less BOPs, with a $1.4\%$ higher accuracy. Furthermore, the BSN offers a great relaxation for a future hardware implementation compared to the alternative approaches to handle normalization layers. Additional results that demonstrate the effectiveness of the proposed framework can be found in the related Appendix~\ref{memory_acc}. 

\begin{table*}[htbp]
	\caption{Comparison of low-precision CNN processors (CIFAR-10 image classification task use case).}
	\begin{center}
		\begin{tabular}{|c|c|c|c|c|c|c|c|c|}
			\hline
			\multicolumn{2}{|c|}{}& Kim et al. & Valavi et al. & Bankman et al. &  Jia et al. &  Cai et al.   & \multicolumn{2}{c|}{\textbf{\textit{This work}}} \\
			\cline{8-9}
			\multicolumn{2}{|c|}{}& \cite{kim_resource-efficient_2021} \textit{TCS'21}&  \cite{valavi_64-tile_2019} \textit{JSSC'19}&  \cite{bankman_always-38_2019} \textit{JSSC' 19} &  
			\cite{jia_bitscale_imc} \textit{JSSC' 20} &  
			\cite{cai_lowbit_rram} \textit{JSSC' 20}   & 
			binary & mixed \\
			\hline
			\multicolumn{2}{|c|}{Weight precision} & 1 & 1 & 1 & 4 & 2 & 1 & 1,2,3 \\
			\hline
			\multicolumn{2}{|c|}{Input Activation precision} & 1 & 1 & 1 & 4 & 4& 1 & 1,2 \\                                    
			\hline
			\multirow{3}{*}{Batch Norm} & Share Level & channel/unit & channel/unit & channel/unit & channel/unit & channel/unit & \multicolumn{2}{c|}{layer} \\
			\cline{2-9}
							& Bias-precision & 6 & 6 & 9 & 6 & 10 & \multicolumn{2}{c|}{0} \\
			\cline{2-9}
							& Implementation & Sign Comp. & Sign Comp. & Addition & Sign Comp. & \makecell{Fused into Kernels} & \multicolumn{2}{c|}{Intrinsic}  \\
			\hline
			\multicolumn{2}{|c|}{On-chip Memory (Mb)} & 14.022 & 2.4 & 2.624 & 29.873   & 18.607  & 0.774 & \textbf{1.073}  \\
			\hline
			\multicolumn{2}{|c|}{MAC$\times$bit ($\times 10^9$)} & 0.617  & 0.158  & 1.007 & 1.847  & 1.253  & 0.125  & \textbf{0.210}  \\
			\hline
			\multicolumn{2}{|c|}{BOPs ($\times 10^9$)} & 0.642 & 0.170  & 1.007 & 7.445    & 5.040   & 0.137 & \textbf{0.287}  \\
			\hline
			\multicolumn{2}{|c|}{Accuracy ($\%$)} &  88.80 & 84.37 & 86.05 & 92.70 & 90.03 & 82.40 & \textbf{87.48} \\
			\hline 
		\end{tabular}
		\label{comparison}
	\end{center}
\end{table*}

\subsection{Full-frame image compression}
In this section we only focus on the reconstruction of images with a VGA resolution ($480\times640$). To this end, each image is divided into $15\times20$ non-overlapping patches of $32\times32$ pixels to apply the NQE compression scheme.   

\textbf{VGA images dataset:} We employ the DIV2K dataset \cite{DIV2K} which includes $800$ images for training and $100$ validation images for testing. In order to meet the target resolution and because the original dataset provides a large variety of image resolutions, the images are cropped accordingly to the target height/width ratio and then resized with a Lanczos kernel with radius of 3. Then we extract all non-overlapping patches of size $32\times32$, obtaining 240k patches for the training of NQE. 

\textbf{NQE training: } The NQE is trained with the PI-PURENET decoder using a batch size of 100, during 60 epochs with the standard BN, and then with 30 epochs after replacing BN by BSN. To train this auto-encoder structure, we used the Mean Square Error (MSE) loss with the Adam optimizer. 

\textbf{PURENET training: } The PURENET decoder is then trained (with a fixed NQE) in 30 epochs with a small batch size of 1 (mainly due to computational resources limitations). To obtain a proper convergence of the decoder, the learning rate is initialized at 0.001 and then rescaled by 0.95 at each epoch after the fifth epoch. We then freeze the Patch-based Upsampling module and fine-tune the Refinement with a batch size of 2 during 30 epochs. The learning rate is also initialized at 0.001 and then rescaled with the same factor 0.95 after the $10^{th}$ epoch.   

\textbf{Comparison with state-of-the-art methods:} 
We compare NQE - PURENET with different methods: JPEG \cite{JPEG} and JPEG2000 \cite{JPEG2000} which are patch-based standard image compression techniques (\ie\, using an entropy coder after a sparsifying transform), one BCS encoding with regularization-based iterative method for decoding consisting in a L1-regularization on a 2D-Daubechies Wavelet Dictionary applied to 3D image gradients (denoted WD-TV3D, inspired from \cite{guicquero_high-order_2015} and giving better results than a basic TV \cite{xiao_alternating_2012}); the end-to-end learning based methods using Recurrent Convolutional AutoEncoder \cite{toderici_full_2017} called RCAE and the Context-Adaptive Entropy Model ( \cite{Lee2019ContextadaptiveEM}) optimized with Peak-Signal-to-Noise Ratio (CAEM-PSNR) or with MultiScale Structural SIMilarity metric \cite{MS_SSIM}(CAEM- MS-SSIM), both dedicated to full-resolution image compression; the Block-Based Decoder (BBD); our NQE with PURENET for patches reconstruction, namely NQE- PI-PURENET. We notice that for WD-TV3D, we apply the same Rademacher matrix to all non-overlapping patches of size 32$\times$32, every measurement is 5-bit uniformly quantized in the range of [-3$\sigma$, 3$\sigma$], where $\sigma = \frac{0.5}{\sqrt{1024}}$ is the estimated standard deviation of the measurements distribution. Besides, CAEM compression rate changes from one image to another, therefore we choose the quality level which allows the average bit-rate close to 0.25bpp.    

\textbf{Image quality evaluation:} The average PSNR and MS-SSIM performances over the 100 test images are reported in Table~\ref{tab:psnr_msssim} while the reconstructed images are displayed in Figure~\ref{image_compression}. It clearly shows that our NQE-PURENET framework delivers better compression quality compared to JPEG and WD-TV3D in both cases patch-based or full-resolution reconstruction, with higher average PSNR/MS-SSIM and better image quality. However, when comparing this method with JPEG2000 and end-to-end learning methods (RCAE, CAEM-PSNR and CAEM- MS-SSIM) which achieve higher PSNR/MS-SSIM thanks to their specific design for this task, we clearly observe the lack of finest details, for example the lion's eye and the castle in the third and the fourth columns. This lack can be easily explained as a direct consequence of the quantized nature of the NQE. 

\textbf{Effect of the Refinement module:} Compared to the BBD and PI-PURENET, the PURENET slightly improves the PSNR but significantly the MS-SSIM, which is easily explainable with respect to the noticeable enhancement in terms of visual quality. We can see the block artifacts in BBD and PI-PURENET, as they do not take into account the surrounding context of each patch. On the contrary, the PURENET successfully renders the smoothness between patches thanks to several convolutional layers which will broaden the information propagation from neighboring patches. This explains why it has MS-SSIM much higher (\ie\, $0.0172$ compared to PI-PURENET and $0.2136$ compared to BBD).
\begin{table}[h]
\renewcommand{\tabcolsep}{1pt}
\centering
\caption{VGA image compression comparison between different methods at the bitrate of $0.25$bpp ($0.2423$bpp for CAEM-PSNR and $0.2459$bpp for CAEM-SSIM).}
\label{tab:psnr_msssim}
\begin{tabular}{|l|*{4}{c|}}\hline
\backslashbox{Method}{Metrics}
&\makebox[5em]{\makecell{Block-based \\ encoder}} &\makebox[5em]{\makecell{Block-based \\ decoder}} &\makebox[5em]{ \makecell{PSNR \\(dB)}}&\makebox[5em]{MS-SSIM}\\
\hline
JPEG & Yes & Yes & 19.82 & 0.7610\\
\hline
JPEG2000 & No & No  & 24.24 & 0.9057 \\
\hline
WD-TV3D & Yes & No  & 19.67 & 0.7255\\
\hline
RCAE & No  & No  & 22.45 & 0.8959\\
\hline
CAEM-PSNR & No & No & 25.33  & 0.9437 \\
\hline
CAEM- MS-SSIM & No & No  &24.19 & 0.9640 \\
\hlineB{2.5}
BBD (Ours) & Yes & Yes & 20.51 & 0.7900 \\ 
\hline
PI-PURENET (Ours) & Yes  & Yes  &  20.58 & 0.7964 \\
\hline
PURENET (Ours) & Yes  & No  & 20.76 & 0.8136 \\
\hline
\end{tabular}
\end{table}

\begin{figure*} [!h]
\setlength\tabcolsep{2pt}%
\begin{tabularx}{\textwidth}{*{6}{C}@{}}
 
   \includegraphics[ width=\linewidth, height=\linewidth, keepaspectratio]{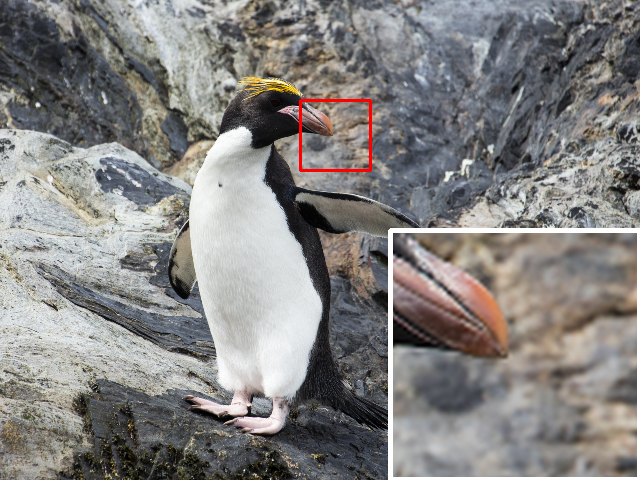} &
   \includegraphics[ width=\linewidth, height=\linewidth, keepaspectratio]{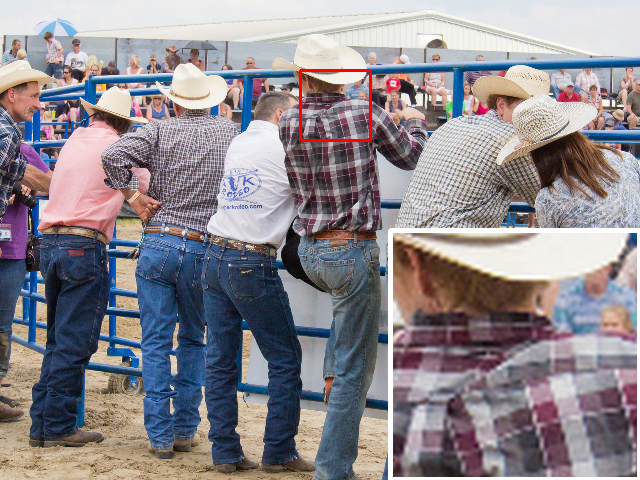} &
   \includegraphics[ width=\linewidth, height=\linewidth, keepaspectratio]{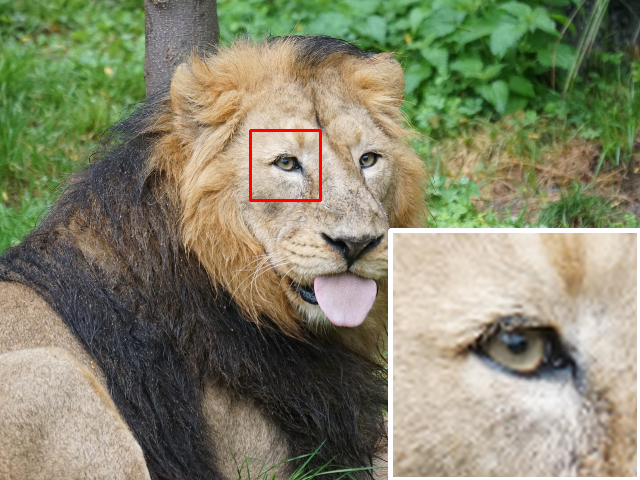} &
   \includegraphics[ width=\linewidth, height=\linewidth, keepaspectratio]{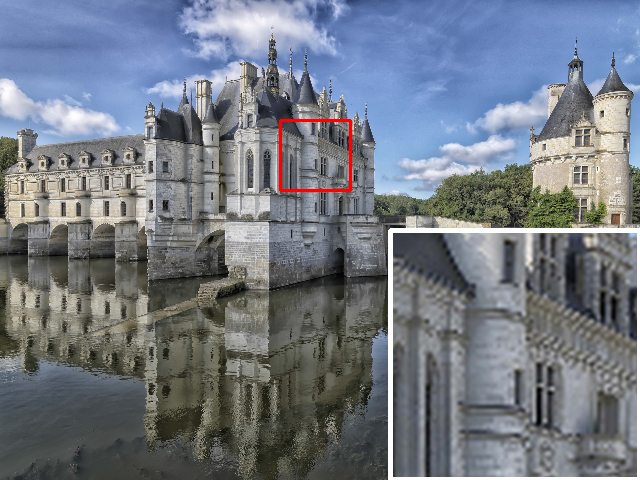} &
   \includegraphics[ width=\linewidth, height=\linewidth, keepaspectratio]{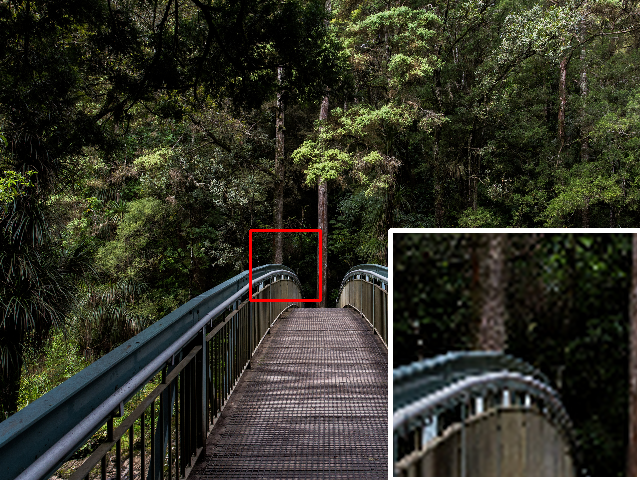} &
   \includegraphics[ width=\linewidth, height=\linewidth, keepaspectratio]{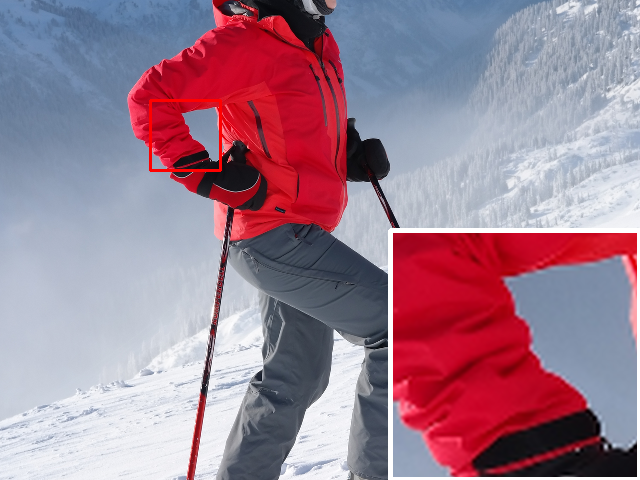} \\
{\small }  & {\small } & {\small } & {\small } & {\small } & {\small } \\ 
\includegraphics[ width=\linewidth, height=\linewidth, keepaspectratio]{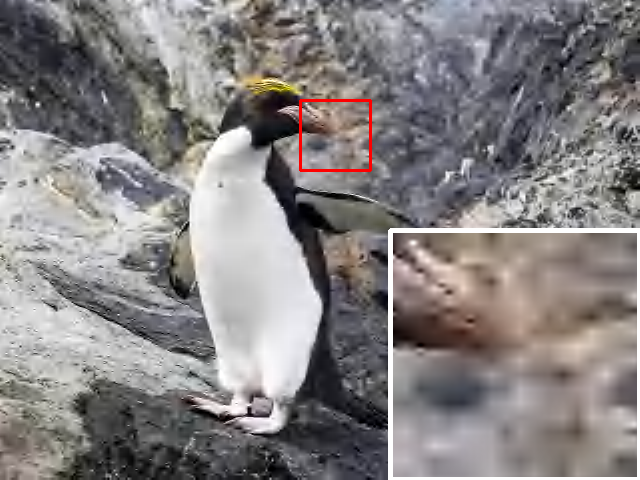} &
   \includegraphics[ width=\linewidth, height=\linewidth, keepaspectratio]{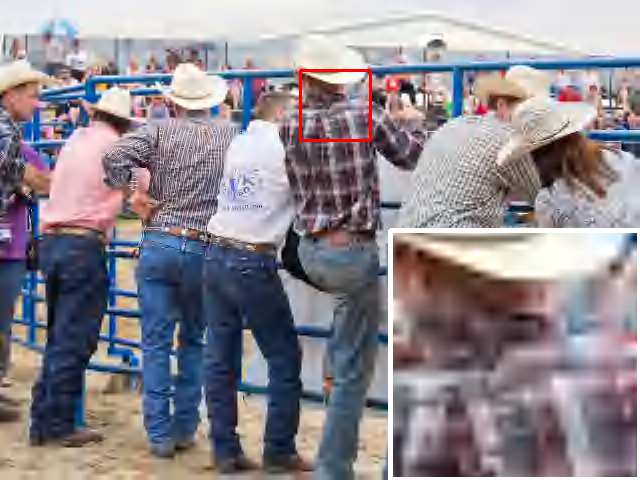} &
   \includegraphics[ width=\linewidth, height=\linewidth, keepaspectratio]{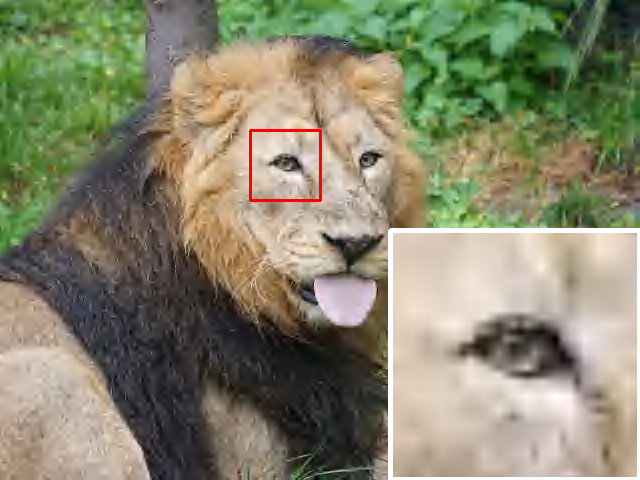} &
   \includegraphics[ width=\linewidth, height=\linewidth, keepaspectratio]{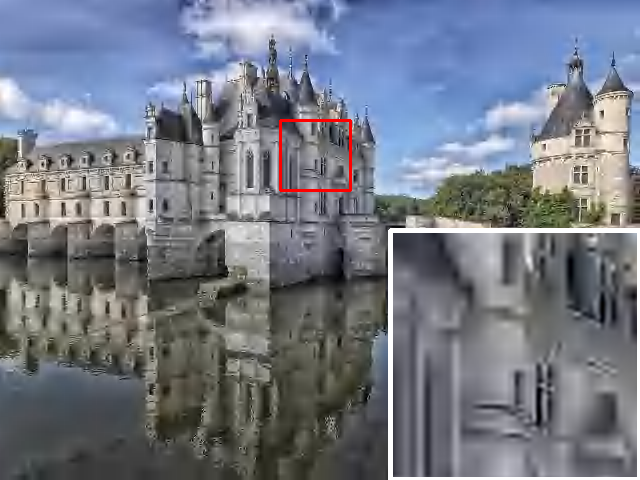} & 
   \includegraphics[ width=\linewidth, height=\linewidth, keepaspectratio]{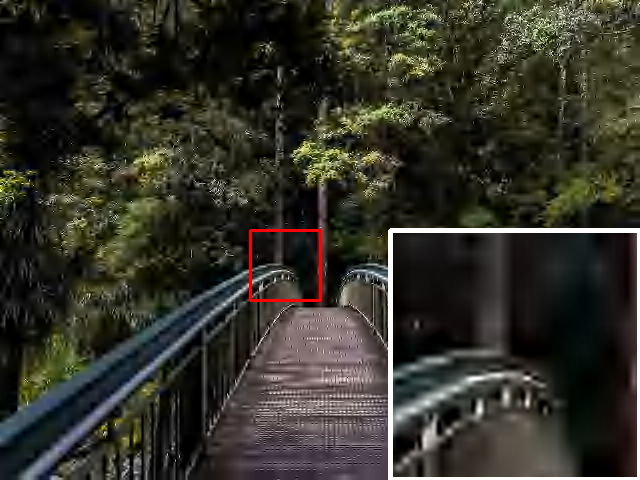}     &
   \includegraphics[ width=\linewidth, height=\linewidth, keepaspectratio]{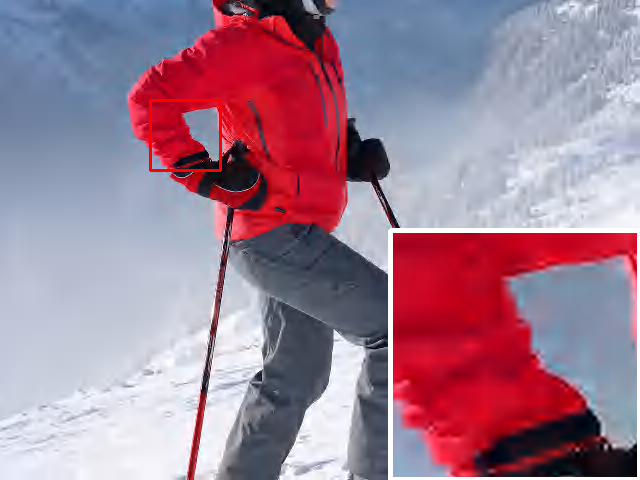}     \\
{\small 23.42/0.9061}  & {\small 22.18/ 0.8884} & {\small 25.54/ 0.8896} & {\small 25.14/ 0.9207} & {\small 20.59/ 0.8303} & {\small 29.78/ 0.9376} \\
\includegraphics[ width=\linewidth, height=\linewidth, keepaspectratio]{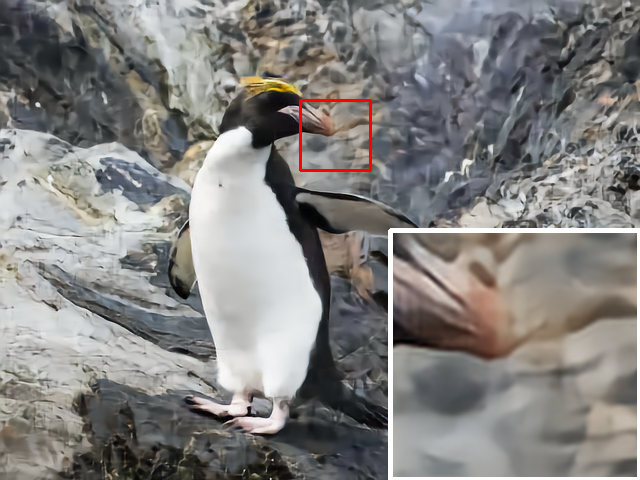} &
   \includegraphics[ width=\linewidth, height=\linewidth, keepaspectratio]{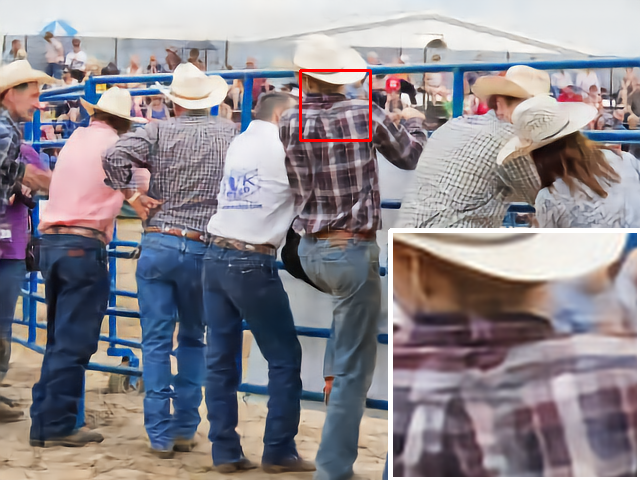} &
   \includegraphics[ width=\linewidth, height=\linewidth, keepaspectratio]{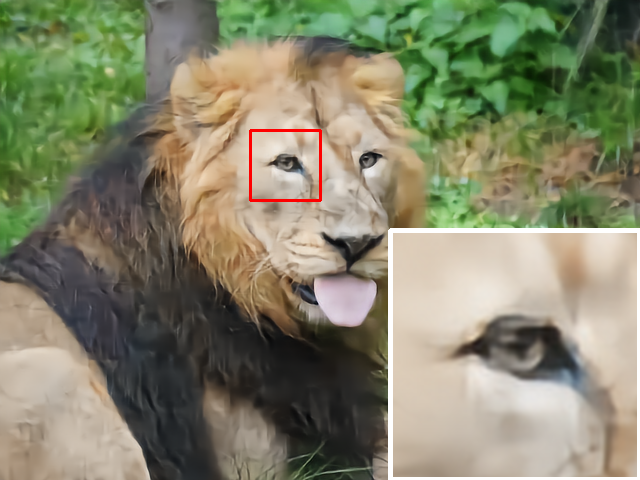} &
   \includegraphics[ width=\linewidth, height=\linewidth, keepaspectratio]{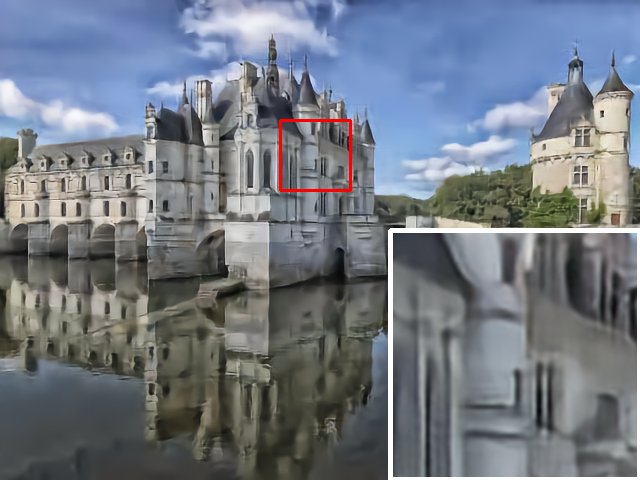} &
   \includegraphics[ width=\linewidth, height=\linewidth, keepaspectratio]{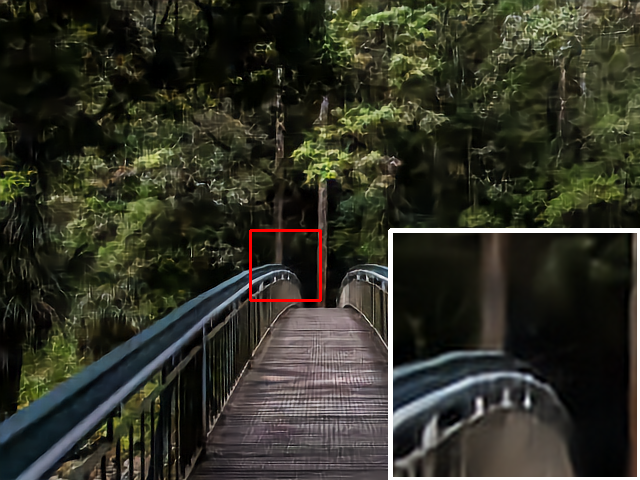}     &
   \includegraphics[ width=\linewidth, height=\linewidth, keepaspectratio]{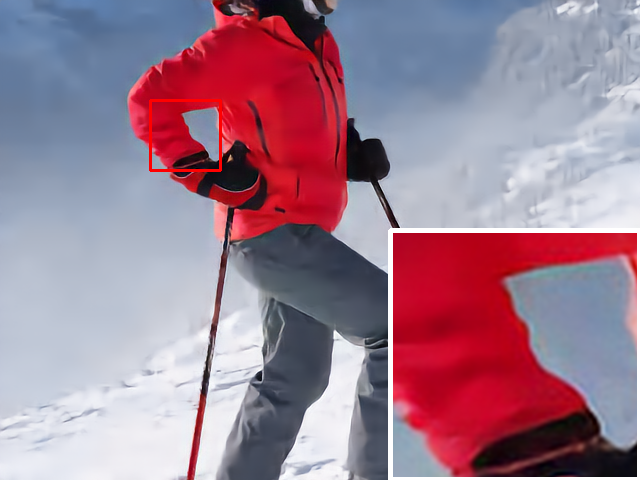}     \\
{\small  24.93/ 0.9414}  & {\small 24.36/ 0.9412} & {\small 26.14/ 0.9028} & {\small 24.43/ 0.9346} & {\small 21.65/ 0.9106} & {\small 29.21/ 0.9362} \\
 \includegraphics[ width=\linewidth, height=\linewidth, keepaspectratio]{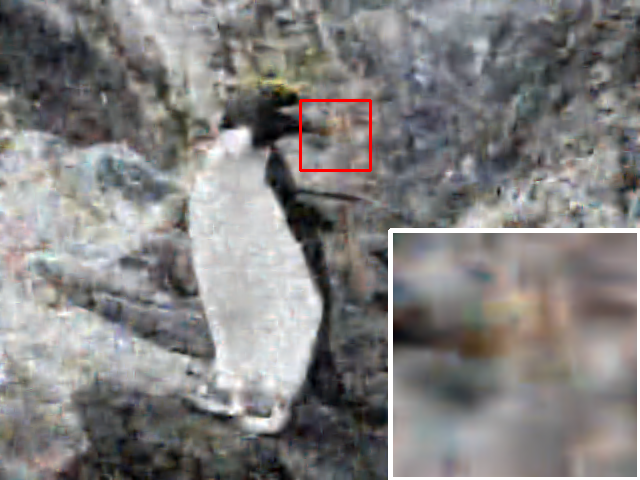} &
   \includegraphics[ width=\linewidth, height=\linewidth, keepaspectratio]{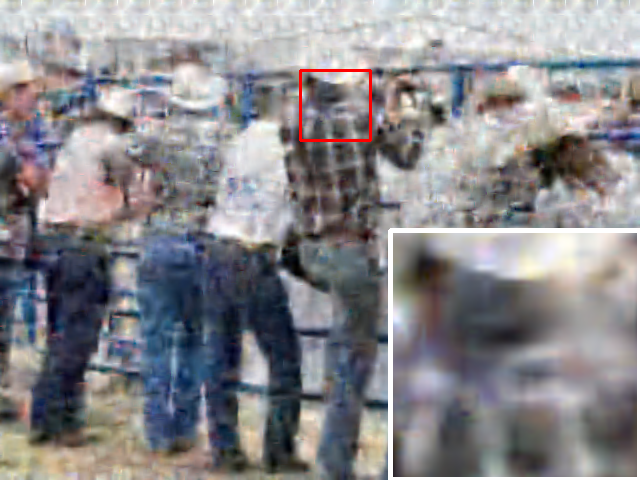} &
   \includegraphics[ width=\linewidth, height=\linewidth, keepaspectratio]{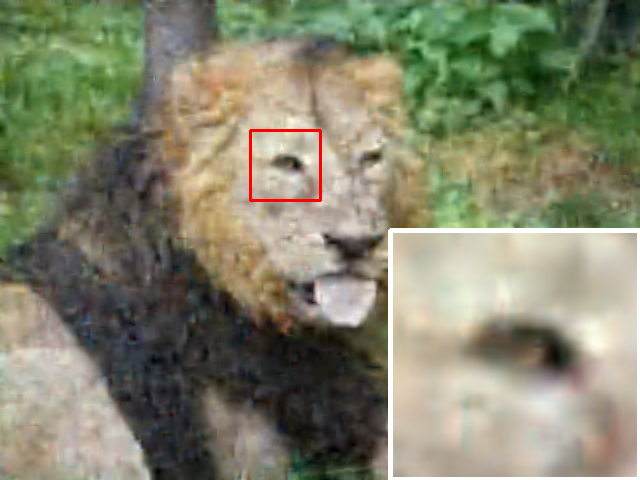} &
   \includegraphics[ width=\linewidth, height=\linewidth, keepaspectratio]{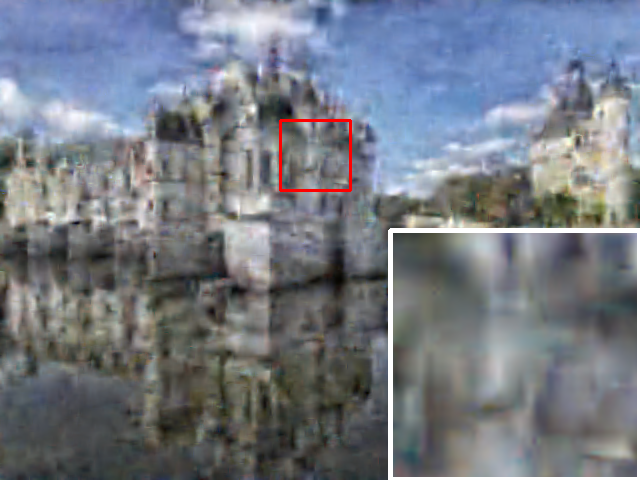} &
   \includegraphics[ width=\linewidth, height=\linewidth, keepaspectratio]{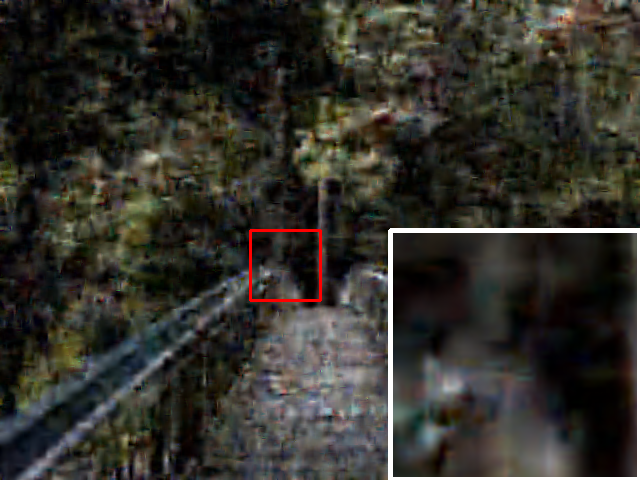}     &
   \includegraphics[ width=\linewidth, height=\linewidth, keepaspectratio]{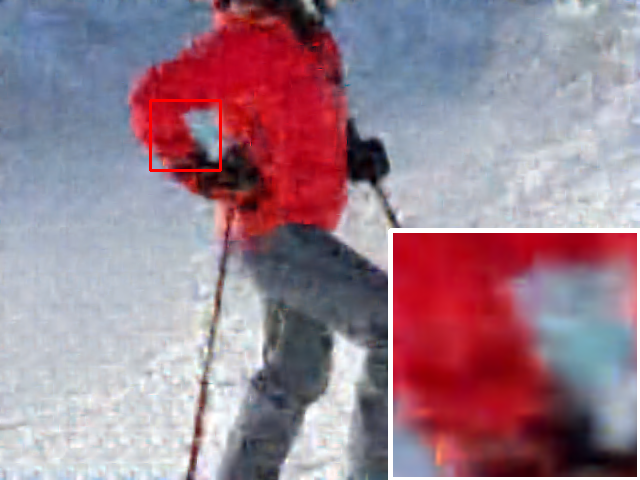} \\
{\small 19.92/ 0.7182}  & {\small 18.52/ 0.7165} & {\small 22.01/ 0.7198} & {\small 21.13/ 0.7680} & {\small 18.18/ 0.6169} & {\small 22.14/ 0.7728} \\
 \includegraphics[ width=\linewidth, height=\linewidth, keepaspectratio]{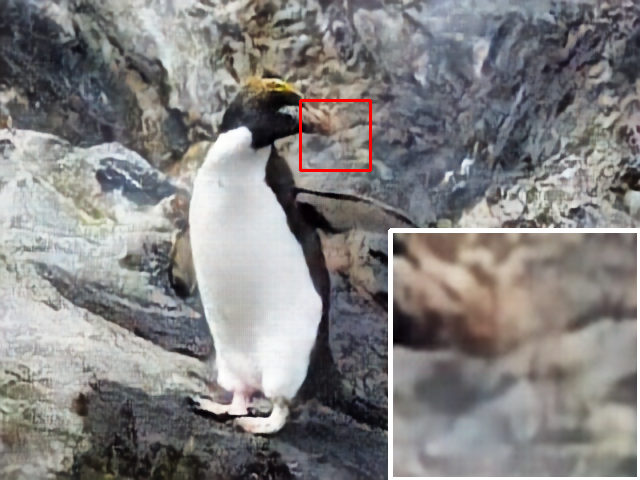} &
   \includegraphics[ width=\linewidth, height=\linewidth, keepaspectratio]{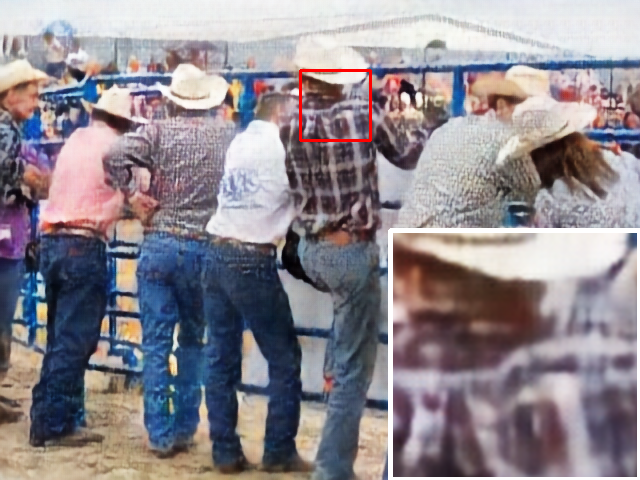} &
   \includegraphics[ width=\linewidth, height=\linewidth, keepaspectratio]{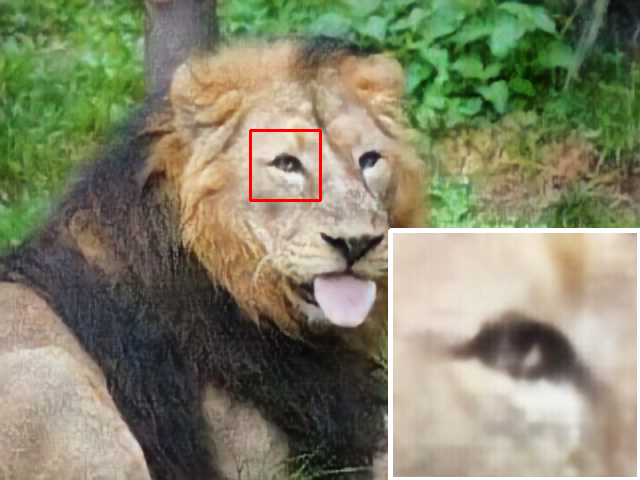} &
   \includegraphics[ width=\linewidth, height=\linewidth, keepaspectratio]{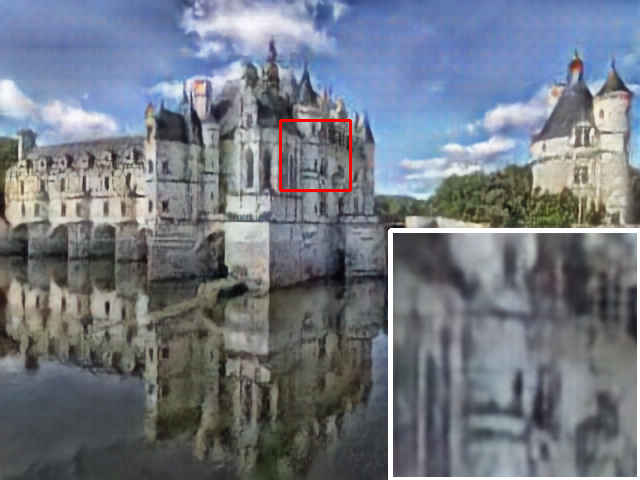} &
   \includegraphics[ width=\linewidth, height=\linewidth, keepaspectratio]{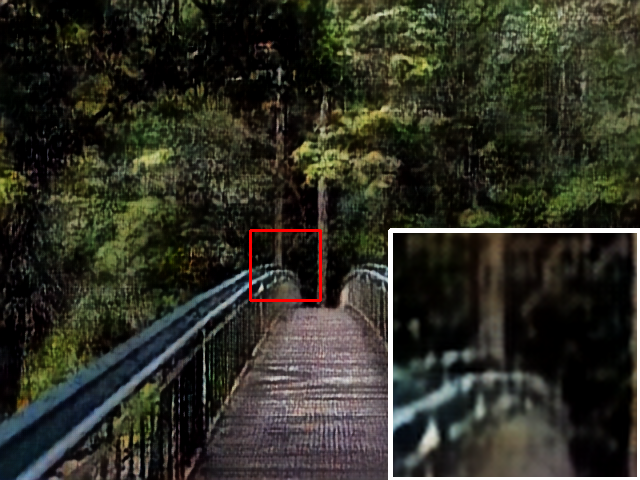}    &
   \includegraphics[ width=\linewidth, height=\linewidth, keepaspectratio]{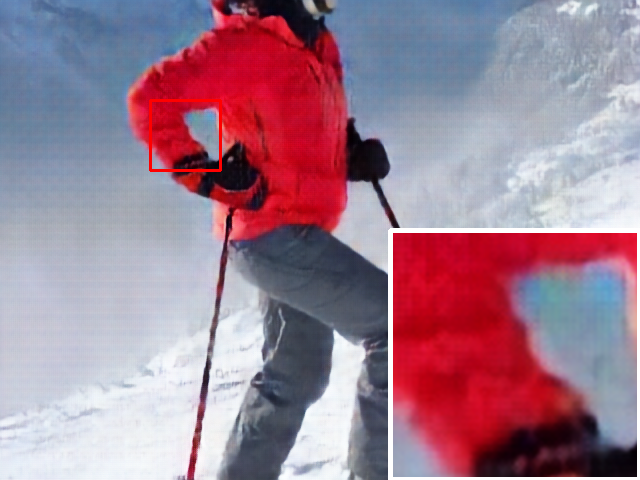}   \\
{\small 22.79/ 0.9082}  & {\small 21.36/ 0.8906} & {\small 24.57/ 0.8877} & {\small 23.65/ 0.9112} & {\small 19.89/ 0.8509} & {\small 26.15/ 0.9119} \\
\includegraphics[ width=\linewidth, height=\linewidth, keepaspectratio]{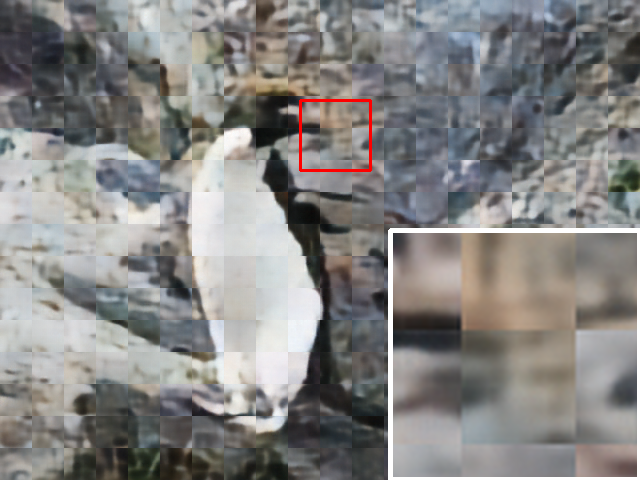} &
   \includegraphics[ width=\linewidth, height=\linewidth, keepaspectratio]{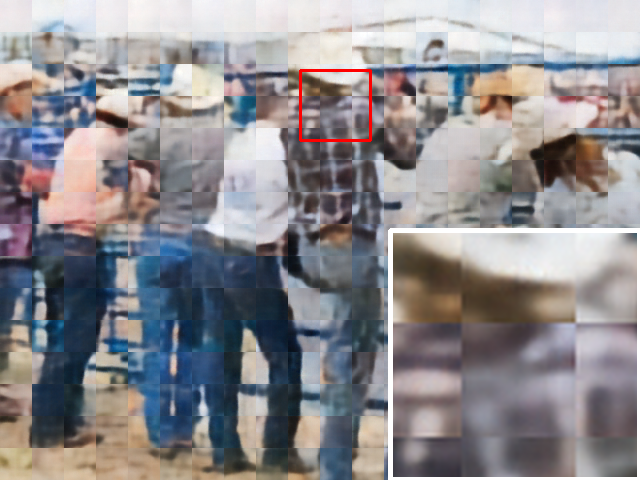} &
   \includegraphics[ width=\linewidth, height=\linewidth, keepaspectratio]{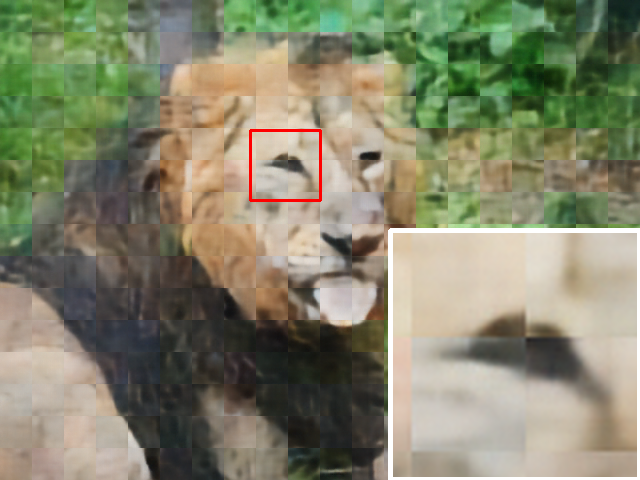} &
   \includegraphics[ width=\linewidth, height=\linewidth, keepaspectratio]{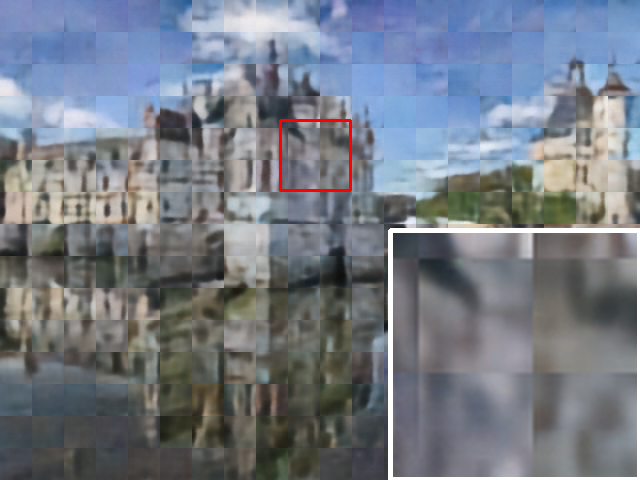} & 
   \includegraphics[ width=\linewidth, height=\linewidth, keepaspectratio]{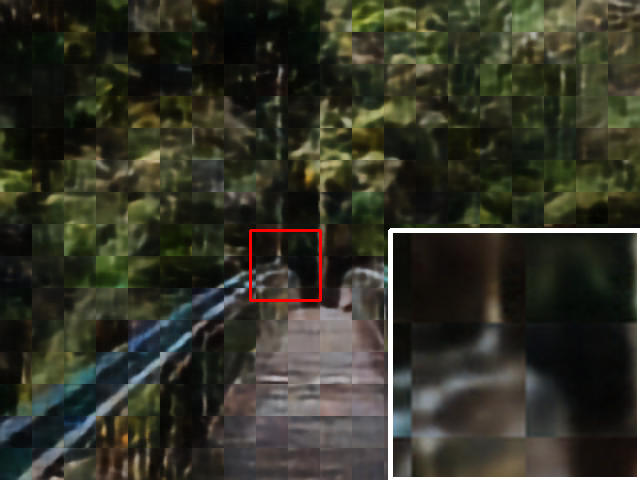}        & 
   \includegraphics[ width=\linewidth, height=\linewidth, keepaspectratio]{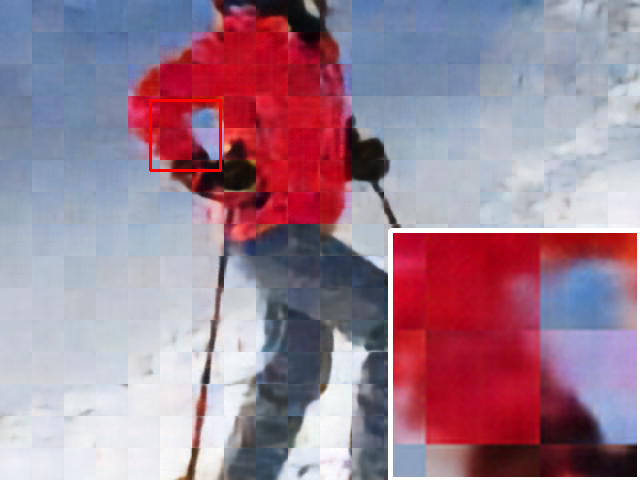}\\
{\small 20.65/ 0.7818}  & {\small 19.49/ 0.7805} & {\small 22.41/ 0.7527} & {\small 21.72/ 0.8018} & {\small 19.02/ 0.7198} & {\small 23.48/ 0.8067} \\
 \includegraphics[ width=\linewidth, height=\linewidth, keepaspectratio]{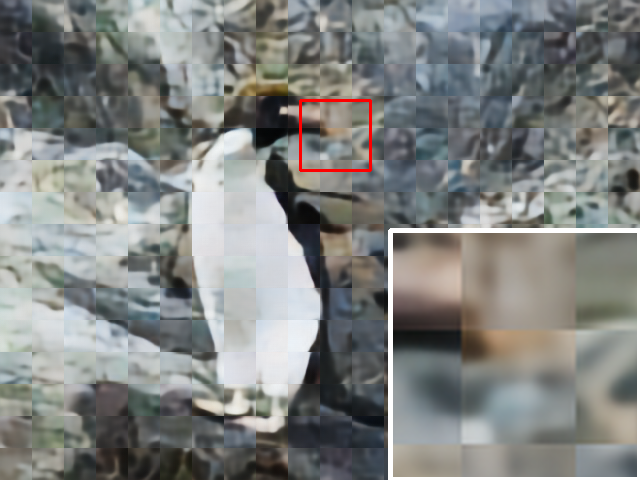} &
   \includegraphics[ width=\linewidth, height=\linewidth, keepaspectratio]{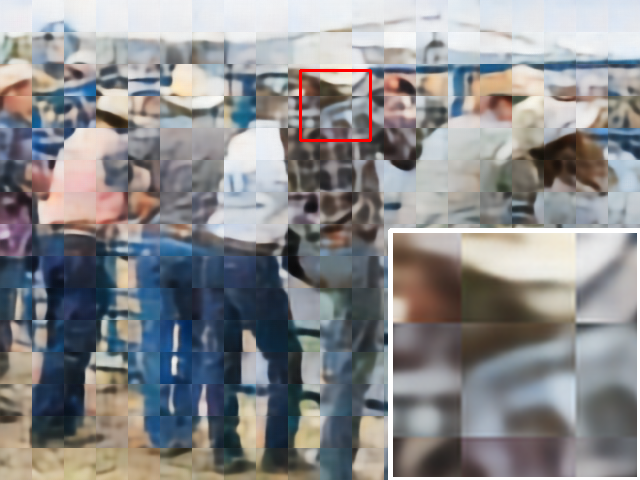} &
   \includegraphics[ width=\linewidth, height=\linewidth, keepaspectratio]{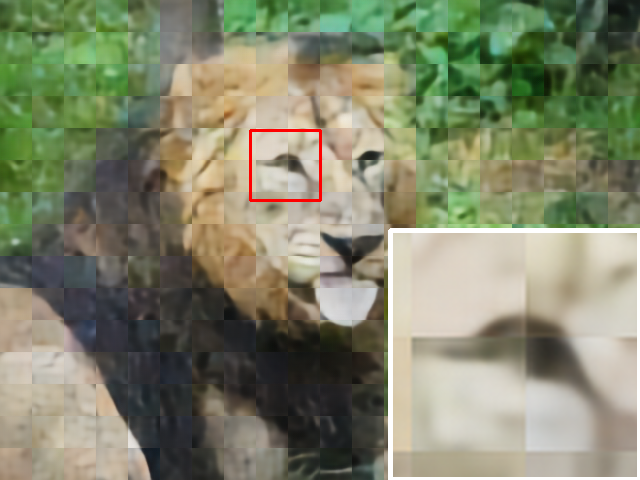} &
   \includegraphics[ width=\linewidth, height=\linewidth, keepaspectratio]{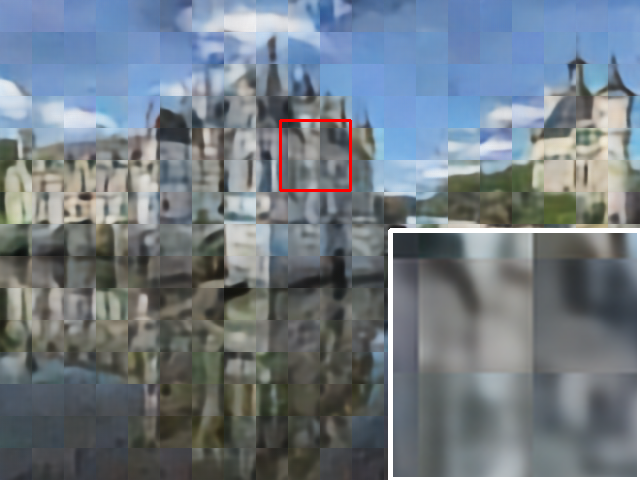} & 
   \includegraphics[ width=\linewidth, height=\linewidth, keepaspectratio]{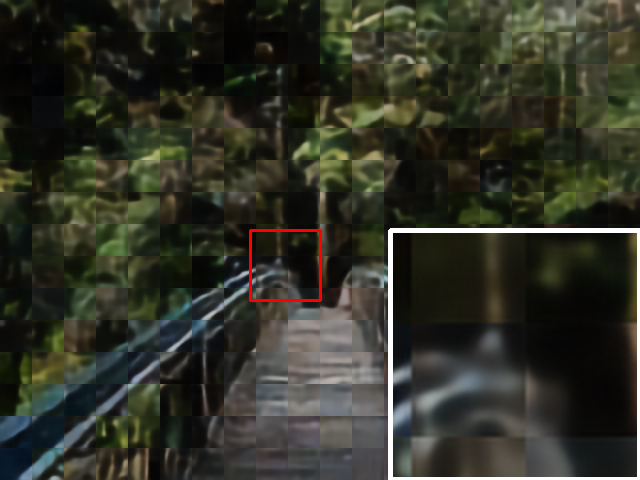}        &
   \includegraphics[ width=\linewidth, height=\linewidth, keepaspectratio]{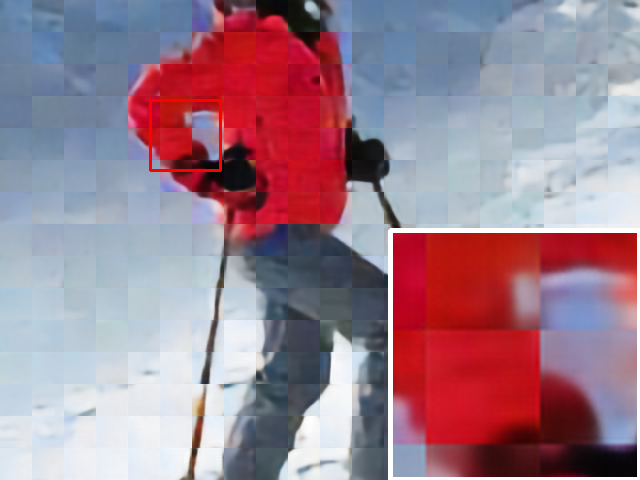} \\
{\small 20.68/ 0.7847}  & {\small 19.57/ 0.7880} & {\small 22.50/ 0.7545} & {\small 21.80/ 0.8089} & {\small 19.09/ 0.7243} & {\small 23.47/ 0.8056} \\
 \includegraphics[ width=\linewidth, height=\linewidth, keepaspectratio]{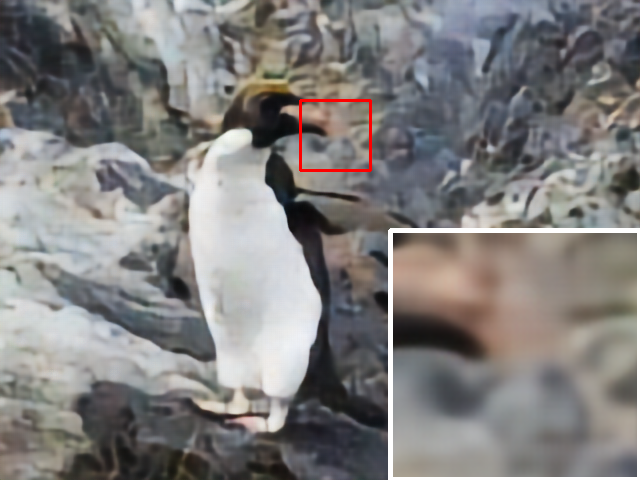} &
   \includegraphics[ width=\linewidth, height=\linewidth, keepaspectratio]{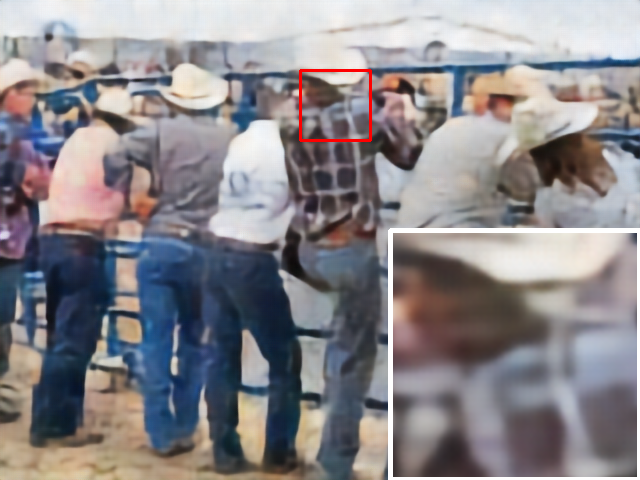} &
   \includegraphics[ width=\linewidth, height=\linewidth, keepaspectratio]{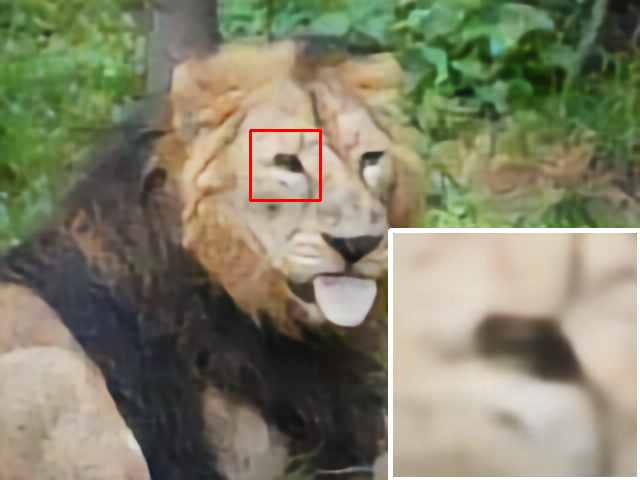} &
   \includegraphics[ width=\linewidth, height=\linewidth, keepaspectratio]{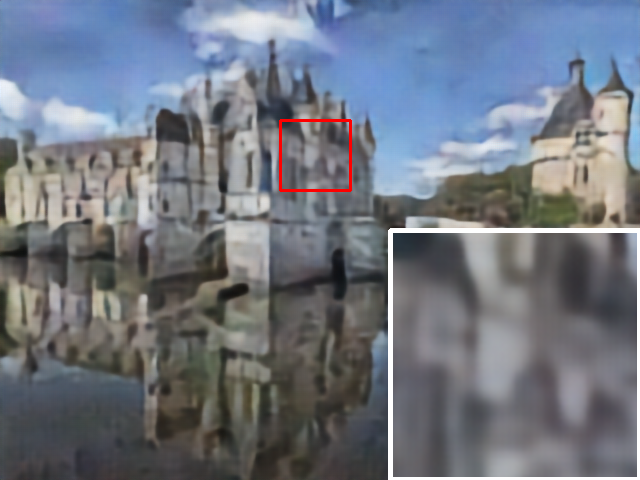} & 
   \includegraphics[ width=\linewidth, height=\linewidth, keepaspectratio]{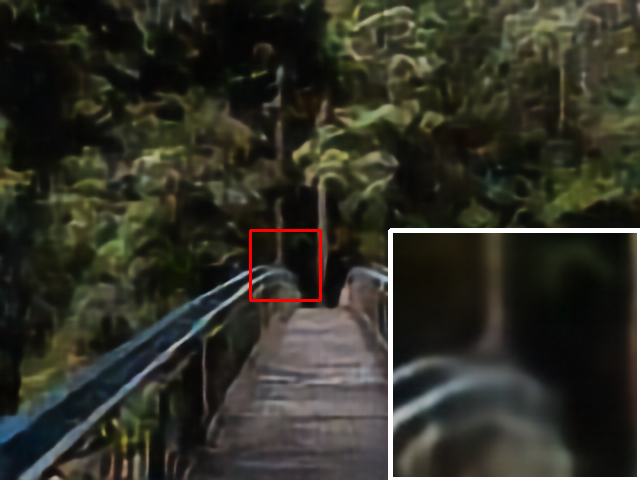}     &
   \includegraphics[ width=\linewidth, height=\linewidth, keepaspectratio]{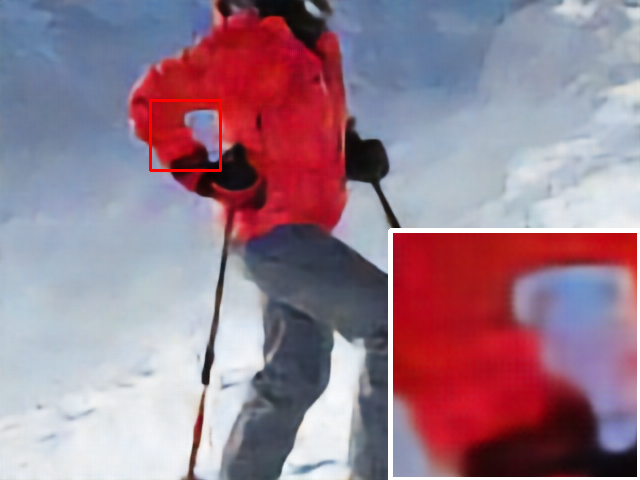}     \\
 {\small 20.79/ 0.7925}  & {\small 19.77/ 0.7997} & {\small 22.71/ 0.7714} & {\small 21.97/ 0.8279} & {\small 19.16/ 0.7291} & {\small 23.67/ 0.8334} \\
\end{tabularx}
\caption{Image compression results at 0.25bpp on some test images of DIV2K validation dataset. From top to bottom: Original image, JPEG2000, CAE-PSNR, WD-TV3D, CAE, our BBD, NQE- PI-PURENET and NQE- Full-Resolution PURENET along with the corresponding \textbf{PSNR}/ \textbf{MS-SSIM} values under each image.} 
\label{image_compression}
\end{figure*}

\begin{table*}[h]
\renewcommand{\arraystretch}{1.1}
	\centering
	\caption{Effect of the bottleneck and its two alternatives to the overall memory and accuracy.}
	\begin{tabular}{|c|c|c|c|c|c|c|c|}
		\hline
		\multirow{2}{*}{{\color{red}F}}  &  \multirow{2}{*}{\makecell{Model memory (Mb) \\ without bottleneck}}  &\multicolumn{2}{c|}{LFC} & \multicolumn{2}{c|}{RCS + FC} & \multicolumn{2}{c|}{\textbf{DWConv + FC (Ours)}} \\
		\cline{3-8}
		&	&  Memory (Mb) & Accuracy (\%) & Memory (Mb) & Accuracy (\%) & Memory (Mb) & Accuracy (\%) \\
		\hline
		 32   &  0.253 & 0.262 & 80.53 &0.016 & 77.98  &0.018  & 79.51  \\
		\hline
		64 & 1.003 & 1.049 & 87.69 &0.066  & 86.33  &0.070 & 87.48 \\
		\hline
		 128 & 3.997 & 4.194  & 90.75 &0.262  & 89.77  &0.270  &  90.15 \\
		\hline	
	\end{tabular}
	\label{tab: bottleneck}
\end{table*}

\textbf{Encoder-Decoder complexity evaluation:} While WD-TV3D makes use of standard linear BCS at the encoder part (very lightweight in terms of hardware), our NQE consists of convolution, bitshift, nonlinear activation. However they all remain hardware-friendly and perform basic feed-forward computations within a limited memory budget of $1$Mb, with the intensive use of low-precision computations. On the contrary, JPEG/JPEG2000 involves more complex algorithmic routines since they take advantage of entropy encoding, requiring non deterministic computational state machines. On the other hand, RCAE uses LSTM cells to deal with various compression rates in a unified model besides an RNN-based entropy coding, so increasing computational needs compared to our approach, in addition to requiring fully floating-point precision. Similarly, CAEM employs much more complex full-precision model combined with entropy coding modules. On the decoder side, WD-TV3D uses power-hungry regularization algorithms while CAE decoder remains computationally complex (mainly due to LSTM cells) whereas JPEG/JPEG2000 decompression is relatively straightforward. In comparison, our PURENET is a floating-point convolutional network with dense skip connections but without any recurrent modules. In conclusion, our Full-Resolution PURENET that corresponds to a simple feed-forward DNN, when combined with NQE, achieves a good complexity-quality trade-off as it can be easily implemented close to the sensor and still allow to properly recovers images, outperforming JPEG and linear BCS compression schemes.

\section{Conclusion}
This paper presents a mixed-precision CNN topology which is compliant with a ASIC design, enabling to perform both low-complexity image classification and embedded patch-based compression. The reported results demonstrate the possible degree of versatility in terms of application use cases for a specific neural network architecture targeting an ASIC design. Numerically speaking, our NQE exhibits a 87.48$\%$ accuracy for CIFAR-10 while requiring $1$Mb of memory and whose weights and activations are quantized with a mixed-precision approach. In addition, Batch Normalization layers are replaced by layer-shared BitShift Normalisations, in order to further ease a possible hardware implementation. In terms of image compression, our PURENET architecture typically deals with patch-based binary coding to perform a collaborative reconstruction, providing images with a relatively high rendering quality at a bitrate of only 0.25bpp. Besides, PSNR and MS-SSIM metrics are better than relevant alternatives such as JPEG and BCS compression schemes. The proposed approach shows a good quality of service versus its computational complexity, especially for an embedded patch-based image compression. Aforementioned results all confirm the advantage of an algorithm/hardware co-design to reach the best trade-off between hardware implementation complexity and algorithmic accuracy.
A direct extension may involve the use of quantized RNN units to extend NQE compression and classification capabilities to frame sequences, not only still images. Another approach may also be using ensemble learning like \cite{9109567} (thanks to the reconfigurability) to enhance the compression quality. In other respects and to improve the performances of the NQE, its topology would also benefit from the use of skip connections so as self-attention mechanisms.

\appendices

\section{Effect of the bottleneck alternative}
To highlight the efficiency of the proposed alternative to the dense layer, we conduct a study on the impact of different types of bottleneck. We compare our proposition with two other settings. The first option is a canonical Fully Connected denoted as \textbf{LFC}, \ie\, a Learnable dense layer with binary weights ($256F^2$ bits). The second option is the same FC layer but with fixed weights (\ie\, not trained) following a zero-mean Rademacher distribution (RCS), followed by a small FC layer of binary weights from $4F$ to $4F$ ($16F^2$ bits), denoted as \textbf{(RCS)$+$FC}. This second option needs 16 less bits to store the weights compared to the LFC, since the Rademacher matrix can be generated on-chip with a pseudo random generator. Table~\ref{tab: bottleneck} reports our results, clearly demonstrating that a large LFC layer of $256F^2$ consumes more than $50\%$ of the overall memory in all cases. When the model size is small, this parameter-heaviness is crucial to improve the model's performance, explaining why the gap between LFC and its two alternatives is more important at $F=32$ (more than $1\%$). On the contrary, its two alternatives contribute only around $6\%$ of the overall memory. The proposed DWConv$+$FC obtains much higher inference accuracy, while increasing by just a tiny amount the memory budget compared to RCS$+$FC ($<10$kb).

\section{Memory vs Accuracy curves}
\label{memory_acc}
\begin{figure}
	\centerline{\includegraphics[scale=0.37]{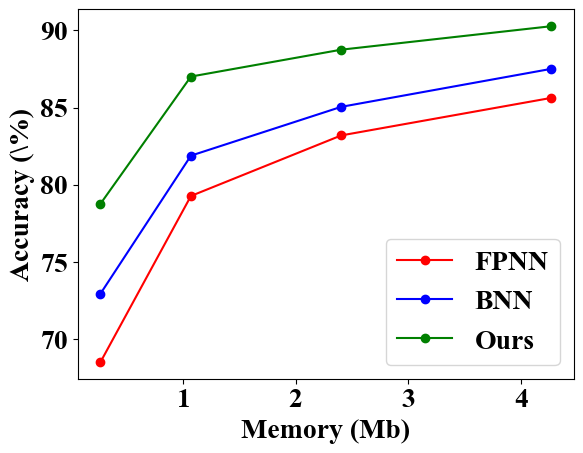}}
	\caption{Memory-accuracy curves of different model's precision (floating-point, binary and ours mixed-precision).} 
	\label{memory_acc_curve}
\end{figure}

Apart from the $1$Mb model, we also report the accuracy for different feature map sizes (Figure~\ref{memory_acc_curve}). It shows that our mixed-precision outperforms FPNN (VGG7 with full-precision) and BNN (VGG7 with binarized weights and input activations) at iso-memory. It even exhibits a large gap, as it provides an efficient design to capture enough discriminant information, compared to the loss due to either a full binarization (BNN) or a ''too tiny size'' (FPNN). 

\newpage
\bibliographystyle{IEEEtran}

\newpage 

\end{document}